\pdfoutput=1

\documentclass[11pt]{article}
\usepackage{placeins}
\usepackage[preprint]{acl}

\usepackage{times}
\usepackage{latexsym}

\usepackage[T1]{fontenc}

\usepackage[utf8]{inputenc}

\usepackage{microtype}

\usepackage{inconsolata}

\usepackage{graphicx}

\usepackage{enumitem}
\usepackage{amsmath}
\usepackage{colortbl}
\definecolor{lightgray}{gray}{0.9}
\definecolor{deepgreen}{rgb}{0,0.5,0}
\usepackage{multirow}
\usepackage{amsfonts}
\usepackage{booktabs}
\usepackage{algorithm}
\usepackage{algorithmic}
\usepackage{supertabular}
\usepackage{longtable}
\usepackage{tabu} 
\usepackage{ragged2e} 
\usepackage{makecell}

\title{MAXS: \underline{M}eta-\underline{A}daptive E\underline{x}ploration with LLM Agent\underline{s}}

\author{Jian Zhang$^{1}$, Zhiyuan Wang$^{1}$, Zhangqi Wang$^{1}$, Yu He$^{1}$\thanks{Corresponding authors}, Haoran Luo$^{2}$,\\
\bf{Li Yuan$^{4}$, Lingling Zhang$^{1}$, Rui Mao$^{2}$, Qika Lin$^{3*}$, Jun Liu$^{1}$}\\
	$^{1}$Xi'an Jiaotong University\;\;
    $^{2}$Nanyang Technological University\\
    $^{3}$National University of Singapore\;\;
    $^{4}$South China University of Technology\\
    \texttt{zhangjian062422@stu.xjtu.edu.cn}, \texttt{heyucs@stu.xjtu.edu.cn}, \texttt{qikalin@foxmail.com}}

\begin{document}
\maketitle

\begin{abstract}

Large Language Model (LLM) Agents exhibit inherent reasoning abilities through the collaboration of multiple tools.  
However, during agent inference, existing methods often suffer from (i) \textit{locally myopic generation}, due to the absence of lookahead, and (ii) \textit{trajectory instability}, where minor early errors can escalate into divergent reasoning paths. These issues make it difficult to balance global effectiveness and computational efficiency.
To address these two issues, we propose \underline{\textbf{m}}eta-\underline{\textbf{a}}daptive e\underline{\textbf{x}}ploration with LLM agent\underline{\textbf{s}} (\textbf{MAXS})\footnote{\url{https://github.com/exoskeletonzj/MAXS}}, a meta-adaptive reasoning framework based on LLM Agents that flexibly integrates tool execution and reasoning planning. 
MAXS employs a lookahead strategy to extend reasoning paths a few steps ahead, estimating the advantage value of tool usage, and combines step consistency variance and inter-step trend slopes to jointly select stable, consistent, and high-value reasoning steps. 
Additionally, we introduce a trajectory convergence mechanism that controls computational cost by halting further rollouts once path consistency is achieved, enabling a balance between resource efficiency and global effectiveness in multi-tool reasoning. We conduct extensive empirical studies across three base models (MiMo-VL-7B, Qwen2.5-VL-7B, Qwen2.5-VL-32B) and five datasets, demonstrating that MAXS consistently outperforms existing methods in both performance and inference efficiency. Further analysis confirms the effectiveness of our lookahead strategy and tool usage.
\end{abstract}

\section{Introduction}
\begin{figure}[t]
\centering
\includegraphics[width=1\columnwidth]{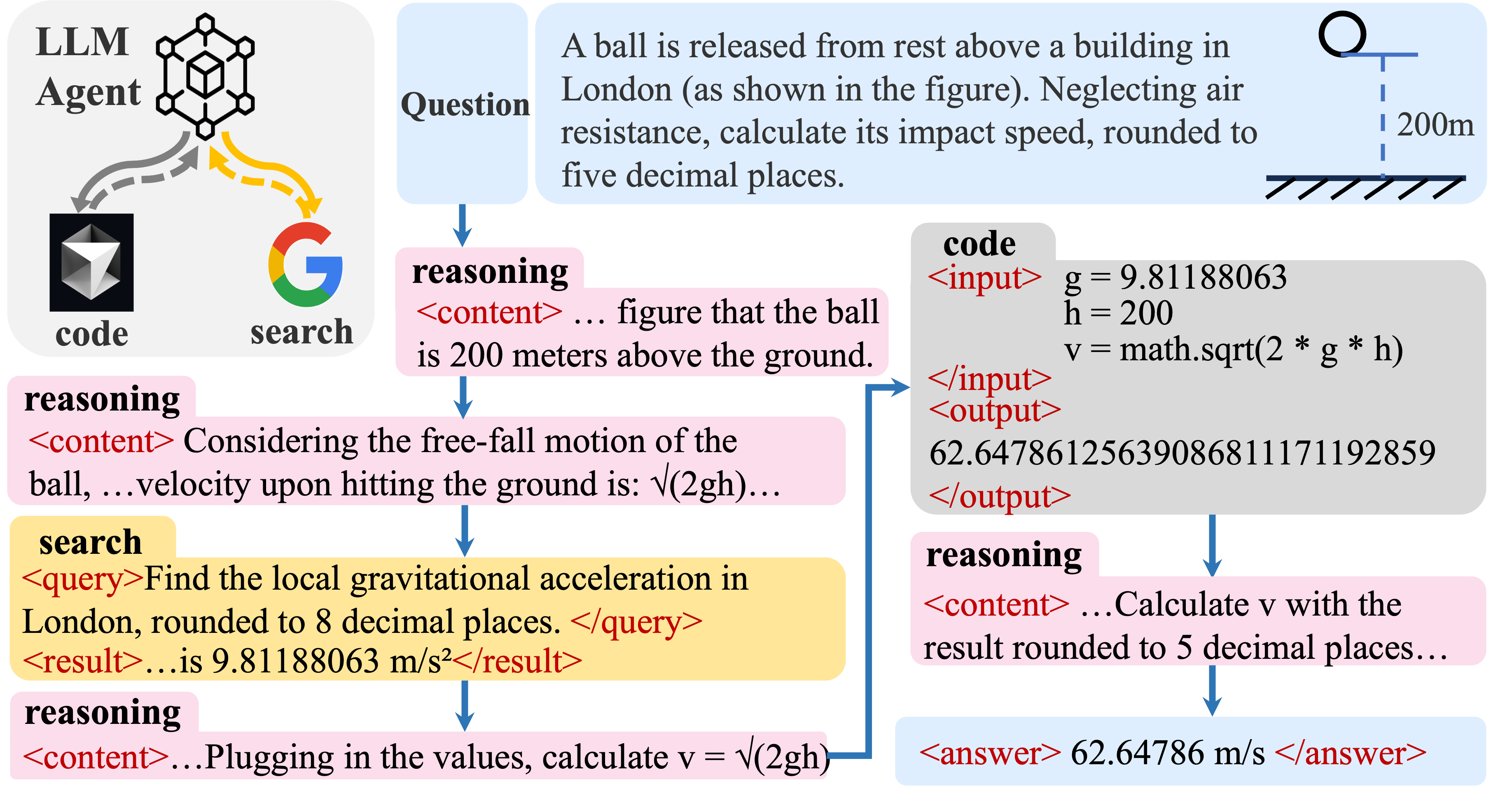}
\caption{An example of LLM Agents solving a task via multi-step reasoning, dynamically leveraging search and code tools to obtain the final answer.}
\label{fig1}
\end{figure}

Large Language Model (LLM) Agents~\cite{huang2024understanding} are built on the backbone of LLM, aiming to leverage tools such as search tools and code tools to assist in the reasoning process.
LLM Agents are widely used in complex problem-solving~\cite{renze2024self}, medical question-answering~\cite{yang2024llm}, search engines~\cite{nie2024hybrid}, and more. Typically, LLM agents generate queries based on reasoning requirements and invoke the search tool to obtain domain-specific knowledge and the latest information, and then use it to obtain the corresponding response~\cite{jin2025search}. LLM Agents use the code tool to generate code based on the reasoning needs, which is then executed by an interpreter to return results for precise calculations~\cite{wang2024executable}. During the reasoning process, LLM Agents appropriately call on the search tool and the code tool to supplement its capabilities and derive the final result, as shown in Figure~\ref{fig1}.

\begin{figure*}[t]
\centering
\includegraphics[width=1\textwidth]{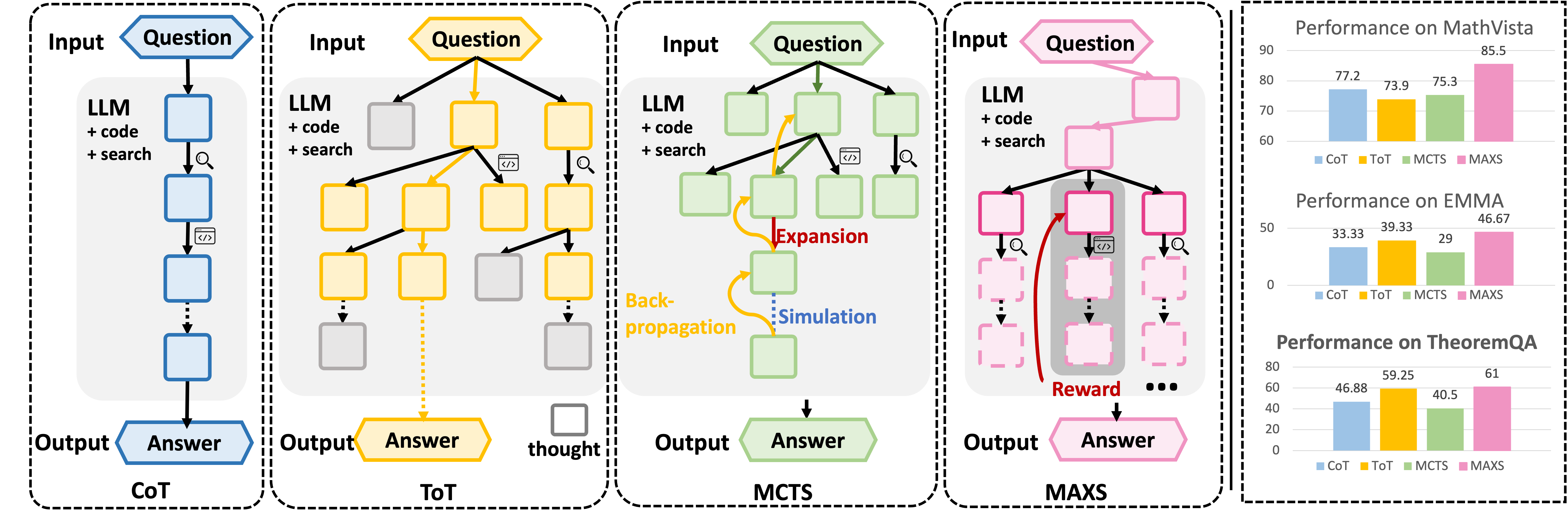} 
\caption{Comparison of test time reasoning strategies. CoT and ToT follow step by step generation with limited foresight, while MCTS conducts global simulation at a higher computational cost. On the right, MAXS uses MiMo-VL-7B-SFT as the backbone and consistently outperforms baseline methods across benchmarks.}
\label{fig2}
\end{figure*}

Various strategies are employed during test-time to improve the efficiency of LLM Agents. As shown in Figure~\ref{fig2}, both \textit{Chain of Thought (CoT)}~\cite{wei2022chain,choi2024embodied} and \textit{Tree of Thought (ToT)}~\cite{yao2023tree,haji2024improving} adopt step-by-step reasoning, following prompt-driven incremental trajectories. In contrast, Monte Carlo Tree Search (MCTS)~\cite{luo2025kbqa,gan2025master} performs global exploration by simulating whole future paths, where each candidate step is evaluated by executing it to completion.

These methods face two major issues. The first is \textbf{locally myopic generation}. Whether using CoT or ToT, both approaches rely on the existing sequence for myopic generation. However, in the context of Agents, crucial factors such as whether a tool should be used, whether its use is appropriate, and whether it brings added value are not reflected in the decision-making process.
The second issue is \textbf{trajectory instability}. Multi-tool reasoning paths are highly sensitive to early decisions, as small errors can accumulate and cause divergence. Tree-based methods like MCTS mitigate this by simulating multiple futures, but at high cost. As shown in Figure ~\ref{fig4}, MCTS often consume approximately one thousand times more tokens to reach similar performance, due to full-path expansion at each step.

To address these issues, we propose \textit{\underline{m}eta-\underline{a}daptive e\underline{x}ploration with LLM agent\underline{s}} (MAXS), a meta-adaptive reasoning framework based on LLM Agents that flexibly integrates tool execution and reasoning planning. MAXS employs a \textbf{lookahead strategy} to extend reasoning paths by a few steps, estimating the potential value of tool usage. It combines step consistency variance and inter-step trend slopes to jointly select stable, consistent, and high-value reasoning steps. Additionally, we introduce a \textit{trajectory convergence} mechanism to control computational costs and improve inference efficiency by halting further rollout once path consistency is achieved. MAXS strikes a balance between resource efficiency and global effectiveness within multi-tool reasoning trajectories.

We conduct extensive empirical studies across five datasets, including \textit{MathVista}~\cite{lu2023mathvista}, \textit{OlympiadBench}~\cite{he2024olympiadbench}, \textit{EMMA}~\cite{hao2025can}, \textit{TheoremQA}~\cite{chen2023theoremqa}, and \textit{MATH}~\cite{hendrycks2021measuring}, 
using three LLM backbones: MiMo-VL-7B~\cite{yue2025mimo}, Qwen2.5-VL-7B~\cite{xu2025qwen2}, and Qwen2.5-VL-32B. 
As shown in the results in Figure~\ref{fig2} and Table~\ref{table:main}, \textit{MAXS} outperforms existing methods in both performance and inference efficiency. Ablation studies further validate the effectiveness of the lookahead strategy and tool usage design. Additional experiments confirm the robustness and adaptability of \textit{MAXS} with multi-tool reasoning trajectories.
The main contributions of this study are threefold:

$\bullet$ We propose a meta-adaptive agent reasoning framework, \textit{MAXS}. To the best of our knowledge, it is the first method to apply \textit{meta-adaptive exploration} during the inference-time of LLM Agents.

$\bullet$ A lookahead-based estimation strategy alleviates both locally myopic generation and trajectory instability by enabling foresighted, value-aware tool selection and promoting stable reasoning paths.

$\bullet$ Extensive experiments across multiple models and datasets demonstrate the effectiveness of \textit{MAXS}, with ablations and further analyses confirming the key role of the lookahead strategy and tool usage design.

\begin{figure*}[t]
\centering
\includegraphics[width=0.85\textwidth]{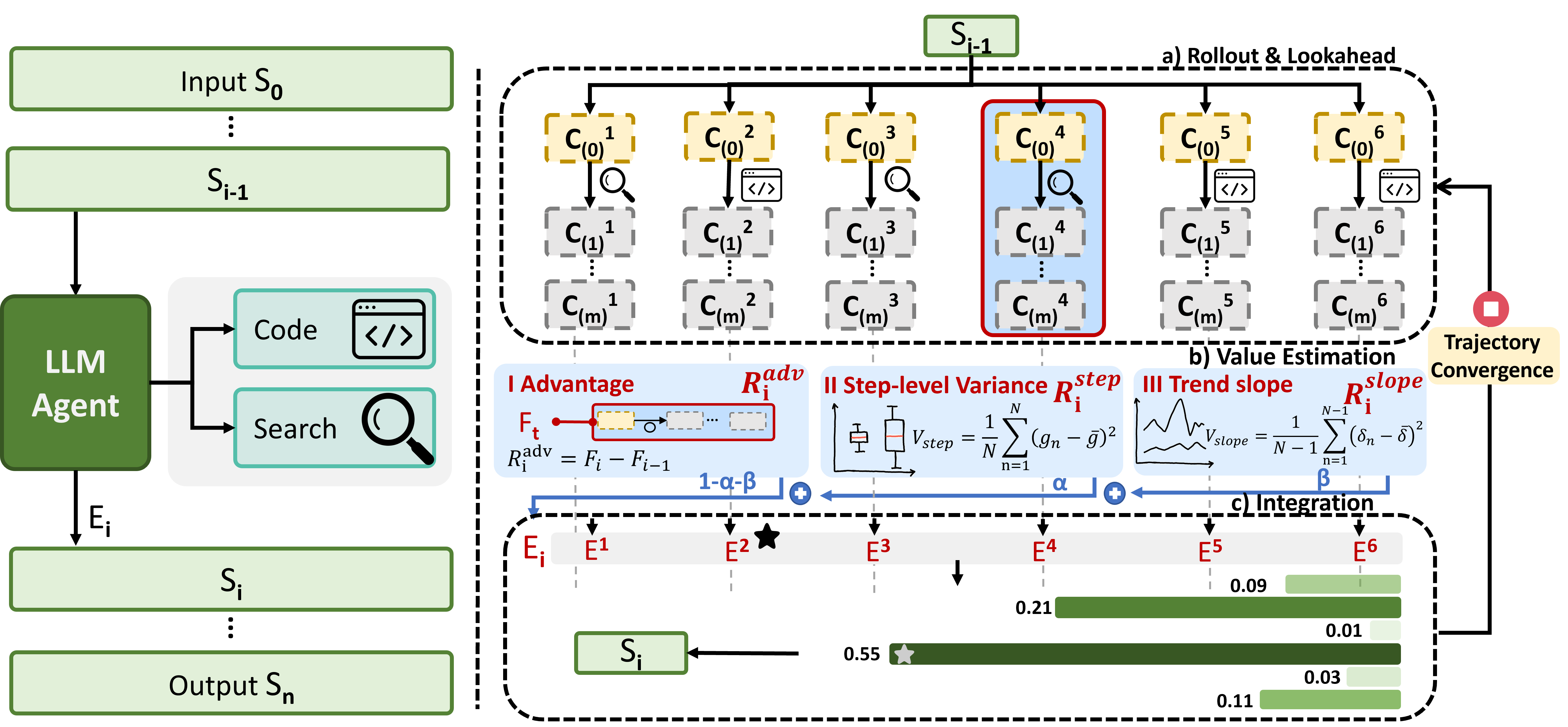} 
\caption{Illustration of the MAXS framework. Left: LLM Agents generates reasoning steps from input $s_0$ to final answer $s_n$. Right: At each step, MAXS performs (a) rollout \& lookahead, (b) value estimation via advantage and two variance scores, and (c) integration. A trajectory convergence mechanism halts rollouts early to improve efficiency.}
\label{fig3}
\end{figure*}

\section{Methodology}

The architecture is illustrated in Figure~\ref{fig3}. In this section, we first introduce the preliminaries of LLM agents-based reasoning. We then present the three key components of MAXS: a lookahead strategy for simulating future steps, a value estimation mechanism for action scoring, and a trajectory convergence module that improves inference efficiency via early rollout termination.

\subsection{Preliminaries}

\textbf{Definition 1: Tool-Augmented Reasoning.}  
LLM Agents reasoning is an iterative process where the agent generates steps $s_i$ based on the reasoning history and input, including the question and prompt $s_0$:
\begin{equation}
s_{i} \sim \pi_{\theta}(\cdot \mid s_0, s_{\leq {i-1}}),
\end{equation}
where $\pi_{\theta}$ is the policy of a pre-trained LLM with parameters $\theta$, and $s_{\leq i}$ denotes all previous reasoning steps.
In tool-augmented settings, the agent can choose to invoke external tools (e.g., search or code) at selected steps $\mathcal{I}_{\text{tool}} \subseteq \{1, \dots, T\}$ to enhance reasoning. The final output $s_n$ is generated by combining the input question $s_0$ with retrieved and computed results:
\begin{equation}
s_n \sim \pi_{\text{final}}\left(s_0;\; \{d_i, r_i\}_{i\in \mathcal{I}_{\text{tool}}} \right).
\end{equation}

\noindent\textbf{Definition 2: Test-Time Strategy.}  
To improve reasoning quality, the agent may apply a selection policy $\mathcal{Q}$ to refine the next step:
\begin{equation}
\hat{s}_i \sim \mathcal{Q}(\cdot \mid s_0, s_{\leq {i-1}}),
\end{equation}
where $\hat{s}_i$ is the selected optimal step, and $\mathcal{Q}$ denotes a test-time strategy such as MCTS.

\noindent\textbf{Definition 3: Search Tool Invocation.}  
At reasoning step $i$, the agent may generate a query to retrieve external knowledge based on input $x$:
\begin{equation}
q_i^{\text{search}} \sim \pi_{\text{search}}(s_0, s_i), \: d_i = \text{Search}(q_i^{\text{search}}).
\end{equation}
The document $d_i$ is used to update the next step.

\noindent\textbf{Definition 4: Code Tool Invocation.}  
At some steps, the agent may also invoke a code tool to perform computation based on the current state and input $x$:
\begin{equation}
c_i \sim \pi_{\text{code}}(s_0, s_i), \: r_i = \text{Exec}(c_i).
\end{equation}
The result $r_i$ is integrated into next reasoning process.

\begin{table*}[htbp]
\centering
\resizebox{\linewidth}{!}{
\begin{tabular}{l|c ccc cccc c c|cc}
\toprule
\multirow{2}{*}{\textbf{Methods}}
& \multirow{2}{*}{\textbf{MathVista}}
& \multicolumn{3}{c}{\textbf{OlympiadBench}} 
& \multicolumn{4}{c}{\textbf{EMMA}} 
& \multirow{2}{*}{\textbf{TheoremQA}} 
& \multirow{2}{*}{\textbf{MATH}} 
& \multirow{2}{*}{\textbf{Avg.}}
& \multirow{2}{*}{\textbf{Tokens}} \\
\cmidrule(lr){3-5} \cmidrule(lr){6-9}
& & math & physics & avg. & Math & Phys. & Chem. & avg. & & & & \\
\midrule
\multicolumn{13}{c}{\textbf{MiMo-VL-7B-SFT}} \\
\midrule
CoT              & \underline{77.20}  & 47.25 & 30.57 & 41.57 & 31.00 & 33.00 & 36.00 & 33.33 & 46.88 & 65.67 & 52.93 & $2.67\times10^7$ \\
ToT              & 73.90  & \underline{48.51} & 32.40 & \underline{43.03} & \underline{39.00} & \underline{39.00} & 40.00 & \underline{39.33} & \underline{59.25} & 69.67 & \underline{57.04} & $6.40\times10^{10}$ \\
MCTS             & 75.30  & 28.98 & 21.83 & 26.55 & 31.00 & 22.00 & 34.00 & 29.00 & 40.50 & 72.67 & 48.80 & $9.91\times10^{10}$ \\
Guided Decoding  & 74.30  & 22.04 & 20.87 & 21.64 & 32.00 & 29.00 & \underline{41.00} & 34.00 & 39.12 & 70.33 & 47.88 & $1.67\times10^8$ \\
$\phi$-Decoding  & 74.80  & 47.86 & \underline{32.79} & 42.73 & 36.00 & 32.00 & \underline{41.00} & 36.33 & 45.75 & \underline{73.00} & 54.52 & $7.66\times10^8$ \\
\midrule
\rowcolor{lightgray}
\textbf{MAXS (ours)}       & \textbf{85.50} & \textbf{52.97}     & \textbf{39.74}     & \textbf{48.47}    & \textbf{47.00} & \textbf{40.00} & \textbf{53.00} & \textbf{46.67} & \textbf{61.00} & \textbf{75.67} & \textbf{63.46}     & $9.86\times10^8$ \\
\midrule
\multicolumn{13}{c}{\textbf{Qwen2.5-VL-7B-Instruct}} \\
\midrule
CoT              & 49.20  & 21.32 & \underline{11.09} & 17.84 & \underline{33.00} & 21.00 & 19.00 & 24.33 & 34.00 & 50.67 & 35.21 & $6.70\times10^6$ \\
ToT              & \underline{52.00}  & 20.03 & 9.48  & 16.44 & 25.00 & 19.00 & \underline{22.00} & 22.00 & 31.00 & 50.00 & 34.29 & $1.37\times10^{10}$ \\
MCTS             & 51.80  & 19.11  & 9.52 & 15.84   & \underline{33.00} & 20.00 & 15.00 & 22.67 & 31.00 & 42.67 & 32.80  & $4.12\times10^{10}$ \\
Guided Decoding  & 44.50  & 25.46 & 10.48 & 20.36 & 32.00 & \underline{27.00} & 16.00 & \underline{25.00} & 34.25 & 53.00 & \underline{35.42} & $1.46\times10^8$ \\
$\phi$-Decoding  & 44.10  & \underline{26.25} & 11.05 & \underline{21.08} & 20.00 & 17.00 & 11.00 & 16.00 & \underline{34.75} & \underline{56.33} & 34.45 & $3.17\times10^8$ \\
\midrule
\rowcolor{lightgray}
\textbf{MAXS (ours)}       & \textbf{56.80} & \textbf{30.49}     & \textbf{15.20}     & \textbf{25.28}     & \textbf{34.00} & \textbf{32.00} & \textbf{30.00} & \textbf{32.33} & \textbf{39.50} & \textbf{60.33} & \textbf{42.85}     & $4.02\times10^8$ \\
\bottomrule
\end{tabular}}
\caption{Main results across five benchmarks using different decoding methods, grouped by models. For OlympiadBench and EMMA, both overall averages and subset performances are reported. The ‘avg.’ column denotes the mean accuracy over MathVista, OlympiadBench(avg.), EMMA (avg.), TheoremQA, and MATH.}

\label{table:main}
\end{table*}

\begin{table}[htbp]
\setlength{\tabcolsep}{4pt}
\centering
\resizebox{\columnwidth}{!}{
\begin{tabular}{l|cccc}
\toprule
\textbf{} & \textbf{Math} & \textbf{Chemistry} & \textbf{Physics} & \textbf{Avg.} \\
\midrule
CoT       & 23.00 & 33.00 & 27.00 & 27.67 \\
ToT       & 25.00 & 22.00 & 24.00 & 23.67 \\
MCTS      & 28.00 & 24.00 & 19.00 & 23.67 \\
Guided Decoding     & \underline{33.00} & 30.00 & 28.00 & 30.33 \\
$\phi$-Decoding     & 31.00 & \underline{35.00} & \underline{33.00} & \underline{33.00} \\
\midrule
\rowcolor{lightgray}
\textbf{MAXS(ours)}          & \textbf{42.00} & \textbf{39.00} & \textbf{37.00} & \textbf{39.33} \\
\bottomrule
\end{tabular}}
\caption{Generalization results on the EMMA dataset using Qwen2.5-VL-32B-Instruct.}
\label{qwen32}
\end{table}

\begin{table*}[htbp]
\setlength{\tabcolsep}{0.9mm}
\centering
\resizebox{\linewidth}{!}{
\begin{tabular}{l|c ccc cccc c c|cc}
\toprule
\multirow{2}{*}{\textbf{Methods}}
& \multirow{2}{*}{\textbf{MathVista}}
& \multicolumn{3}{c}{\textbf{OlympiadBench}} 
& \multicolumn{4}{c}{\textbf{EMMA}} 
& \multirow{2}{*}{\textbf{TheoremQA}} 
& \multirow{2}{*}{\textbf{MATH}} 
& \multirow{2}{*}{\textbf{Avg.}}
& \multirow{2}{*}{\textbf{Tokens}} \\
\cmidrule(lr){3-5} \cmidrule(lr){6-9}
& & math & physics & avg. & Math & Phys. & Chem. & avg. & & & & \\
\midrule
\multicolumn{13}{c}{\textbf{MiMo-VL-7B-SFT}} \\
\midrule
\rowcolor{lightgray}
\textbf{MAXS (ours)}       & \textbf{85.50} & \textbf{52.97}     & \textbf{39.74}     & \textbf{48.47}    & \textbf{47.00} & \textbf{40.00} & \textbf{53.00} & \textbf{46.67} & \textbf{61.00} & \textbf{75.67} & \textbf{63.46}     & $9.86\times10^8$ \\

\hspace{1em}\textit{w/o $lookahead$}      & 78.20  & 49.12  & 30.96  & 42.94  & 42.00  & 36.00  & 49.00  & 42.33  & 58.38  & 70.67  & 58.50  & $2.44\times10^8$ \\
\hspace{1em}\textit{w/o $score_{adv}$}    & 81.60  & 51.74  & 36.68  & 46.61  & 43.00  & 38.00  & 51.00  & 44.00  & 59.25  & 73.33  & 60.96  & $9.88\times10^8$ \\
\hspace{1em}\textit{w/o $score_{step}$}   & 82.40  & 51.15  & 37.12  & 46.37  & 44.00  & 38.00  & 51.00  & 44.33  & 59.63  & 74.00  & 61.35  & $8.32\times10^8$ \\
\hspace{1em}\textit{w/o $score_{slope}$}  & 84.10  & 52.34  & 38.21  & 47.53  & 45.00  & 38.00  & 52.00  & 45.00  & 60.75  & 74.67  & 62.41  & $8.92\times10^8$ \\
\hspace{1em}\textit{w/o $T.C.$}           & 85.10  & 52.41  & 39.04  & 47.86  & 47.00  & 39.00  & 52.00  & 46.00  & 60.88  & 75.33  & 63.03  & $9.95\times10^8$ \\

\midrule
\multicolumn{13}{c}{\textbf{Qwen2.5-VL-7B-Instruct}} \\
\midrule
\rowcolor{lightgray}
\textbf{MAXS (ours)}       & \textbf{56.80} & \textbf{30.49}     & \textbf{15.20}     & \textbf{25.28}     & \textbf{34.00} & \textbf{32.00} & \textbf{30.00} & \textbf{32.33} & \textbf{39.50} & \textbf{60.33} & \textbf{42.85}     & $4.02\times10^8$ \\

\hspace{1em}\textit{w/o $lookahead$}  & 46.30 & 23.46 & 10.17 & 18.94 & 24.00 & 23.00 & 22.00 & 23.00 & 28.50 & 50.33 & 33.41 & $1.76\times10^8$ \\
\hspace{1em}\textit{w/o $score_{adv}$}  & 48.10 & 27.96 & 12.45 & 22.68 & 29.00 & 26.00 & 25.00 & 26.67 & 33.25 & 54.00 & 36.94 & $4.01\times10^8$ \\
\hspace{1em}\textit{w/o $score_{step}$} & 50.40 & 28.41 & 12.71 & 23.07 & 28.00 & 26.00 & 25.00 & 26.33 & 33.88 & 54.67 & 37.67 & $3.87\times10^8$ \\
\hspace{1em}\textit{w/o $score_{slope}$} & 53.10 & 28.77 & 13.14 & 23.45 & 29.00 & 27.00 & 26.00 & 27.33 & 34.75 & 55.33 & 38.79 & $3.97\times10^8$ \\
\hspace{1em}\textit{w/o $T.C.$} & 55.00 & 30.19 & 14.98 & 25.01 & 32.00 & 31.00 & 29.00 & 30.67 & 38.63 & 58.67 & 41.60 & $4.08\times10^8$ \\

\bottomrule
\end{tabular}}
\caption{Ablation results on different backbones. We individually ablate the lookahead module, three value estimation scores, and the trajectory convergence (T.C.) mechanism.  \textit{w/o} denotes experiments conducted without the specified module.}
\label{table:ablation}
\end{table*}

\subsection{Lookahead Strategy}

To mitigate the issue of locally myopic generation, we adopt lookahead via a rollout process. This approach evaluates the current step $s_i$ and future steps $s_{>i}$ to determine the most optimal decision. The lookahead process is defined as:

\begin{equation}
\hat{s}_i \sim \pi_{\theta}(s_i \mid s_0, s_{<i}, s_{>i}),
\end{equation}
where $s_i$ is the current reasoning state, $s_0$ represents the input question and prompt, and $s_{>i}$ includes future steps to be evaluated.

According to the Bellman Optimality Principle~\cite{barron1989bellman}, the value of future steps $R(s_{>i})$ can be recursively estimated as:

\begin{equation}
R(s_0, s \leq i, s > i) = \mathbb{E}\left[\sum_{k=1}^{K} \gamma^{k-1} R(s_{i+k}) \mid s \right],
\end{equation}



where $\gamma$ is the discount factor for future steps, $K$ is the maximum number of steps in the lookahead, and $s$ is the whole steps. This allows us to incorporate future trajectory values into the decision-making process.

\textit{\textbf{Proposition 1 (Bellman Recursion).}  
The optimal action at step \(i\) obeys  
\(\hat{s}_i=\arg\max_{s_i}\bigl[R(s_i*\gamma\,\mathbb{E}_{s_{>i}}V^{\!*}(s_{>i})\bigr]\),  
hence the sequence’s optimum is obtained by recursively combining current utility with the future optimal value.}

The detailed derivation can be found in Appendix~\ref{A.1}. Finally, the current step is selected based on the estimated future values $R(s_{>i})$ as:

\begin{equation}
\hat{s}_i \sim \pi_{\theta}(s_i \mid s_0, s_{<i})\, e^{\frac{R(s_0, s \leq i, s > i)}{\tau}},
\label{equal8}
\end{equation}

where $\tau$ controls the diversity of the generated steps. 
The complete algorithm and decoding pipeline are presented in Appendix~\ref{decoding}.

\subsection{Value Estimation}

To address trajectory instability, a composite value function evaluates candidate reasoning trajectories, incorporating advantage score, step-level variance, and slope-level variance to promote stable and consistent reasoning.

\paragraph{(1) Advantage Score.}  

We adopt beam search to maintain \( K \) candidate paths. At each decoding step \( i \), for each path, we perform \( M \) independent stochastic rollouts to simulate possible future trajectories and evaluate the expected lookahead return~\cite{xu-etal-2025-genius}.
Let \( F_i \) be the foresight probability at step \( i \) under the extended rollout:
\begin{equation}
F_i = \pi_\theta(s_{>i} \mid s_0, s_{\leq i}),
\end{equation}
where \( s_{>i} \) denotes the future \( N \) steps after \( i \). We define the global advantage as the relative improvement over the previous step:
\begin{equation}
A_i = F_i - F_{i-1}, \quad R^{\text{adv}}_i = \exp\left(\frac{A_i}{\tau}\right),
\end{equation}
where \( \tau \) is a temperature parameter controlling sensitivity. \( R^{\text{adv}}_i \) reflects the progress gained by choosing \( s_i \).

\paragraph{(2) Step-Level Variance.}
Inspired by Lyapunov stability theory~\cite{shevitz2002lyapunov}, we interpret the lookahead trajectory as a discrete-time dynamical system. Let \( g_n \) denote the log-probability of the \( n \)-th step in the lookahead segment \( s_{>i} \), and define its mean over a rollout of length \( N \) as \( \bar{g} = \frac{1}{N} \sum_{n=1}^{N} g_n \), and its variance as:
\begin{equation}
V_{\text{step}} = \frac{1}{N} \sum_{n=1}^{N} (g_n - \bar{g})^2.
\end{equation}

Lower \( V_{\text{step}} \) reflects bounded fluctuation across future steps, indicating that the trajectory remains stable and resists erratic deviations, akin to Lyapunov-stable behavior. Accordingly, we define the step consistency reward as \( R^{\text{step}}_i = \exp\left(-\frac{V_{\text{step}}}{\tau}\right) \), where \( \tau \) is a temperature parameter controlling sensitivity.

\textit{\textbf{Proposition 2 (Deviation Bound).}  
If \(V_{\text{step}}\le\varepsilon\), then \(|g_n-\bar{g}|\le\sqrt{N\varepsilon}\) for every \(n\). Bounding \(V_{\text{step}}\) therefore constrains state fluctuations and yields Lyapunov‑like stability.}

The detailed derivation can be found in Appendix~\ref{A.2}. This variance serves as a regularizer to favor smoother forward reasoning paths.

\paragraph{(3) Slope-Level Variance.}
Inspired by Lipschitz continuity in mathematical analysis~\cite{heinonen2005lectures}, we measure the directional smoothness of the lookahead trajectory by evaluating local slope variations. We define the first-order difference \( \delta_n = g_{n+1} - g_n \). The average slope over a rollout of length \( N \) is
\( \bar{\delta} = \frac{1}{N-1} \sum_{n=1}^{N-1} \delta_n \),
and its variance is given by:
\begin{equation}
V_{\text{slope}} = \frac{1}{N-1} \sum_{n=1}^{N-1} (\delta_n - \bar{\delta})^2.
\end{equation}

Lower \( V_{\text{slope}} \) implies the trajectory's local increments are uniformly bounded, resembling Lipschitz-continuous behavior that avoids abrupt changes. Accordingly, we define the slope consistency reward as \( R^{\text{slope}}_i = \exp\left(-\frac{V_{\text{slope}}}{\tau}\right) \), where \( \tau \) controls sensitivity to local oscillations.

\textit{\textbf{Proposition 3 (Lipschitz Bound).}  
If \(V_{\text{slope}}\le\varepsilon\), then for all \(m,n\) we have \(|g_m-g_n|\le\sqrt{(N-1)\varepsilon}\,|m-n|\). Hence bounding \(V_{\text{slope}}\) limits worst‑case jumps and enforces Lipschitz‑like smoothness.}

The detailed derivation can be found in Appendix~\ref{A.3}. This reward encourages the model to prefer directionally coherent forward reasoning paths.

\paragraph{Combining Multiple Rewards.}
We combine the normalized scores of advantage, consistency, and slope into a unified reward:
\begin{multline}
R(s_0, s_{\leq i}, s_{>i})
= (1-\alpha-\beta)\cdot \operatorname{Norm}(R^{\text{adv}}_i) \\
+ \alpha\cdot \operatorname{Norm}(R^{\text{step}}_i)
+ \beta\cdot \operatorname{Norm}(R^{\text{slope}}_i),
\label{eq:joint_reward}
\end{multline}

where each component is temperature-scaled and normalized by
$
\text{Norm}(R_i) = \frac{\exp(R_i / \tau)}{\sum_{j} \exp(R_j / \tau)},
$
with \(\tau = 0.6\).

Replacing this formulation of \(R\) into Eq.~\ref{equal8}, the objective becomes sampling from the joint distribution that captures advantage, consistency, and directional smoothness. 

\subsection{Trajectory Convergence}

To reduce computation and improve inference efficiency, we monitor the variance of candidate rewards \( R(s_0, s_{\leq i}, s_{>i}) \) at each step. Once the variance falls below a threshold \( \delta \), we stop rollout and resume auto-regressive decoding.
Let \( \mathcal{R}_i = \{R^{(k)}(s_0, s_{\leq i}^{(k)}, s_{>i}^{(k)})\}_{k=1}^K \). The early stopping condition is:
\begin{equation}
\text{Var}(\mathcal{R}_i) \leq \delta.
\label{eq:early_stop}
\end{equation}

We terminate rollout at step \( i \) and resume decoding under the auto-regressive process.
For all experiments, we set the convergence threshold \( \delta = 0.002 \) to balance efficiency and stability.

\begin{figure}[t]
\centering
\includegraphics[width=\columnwidth]{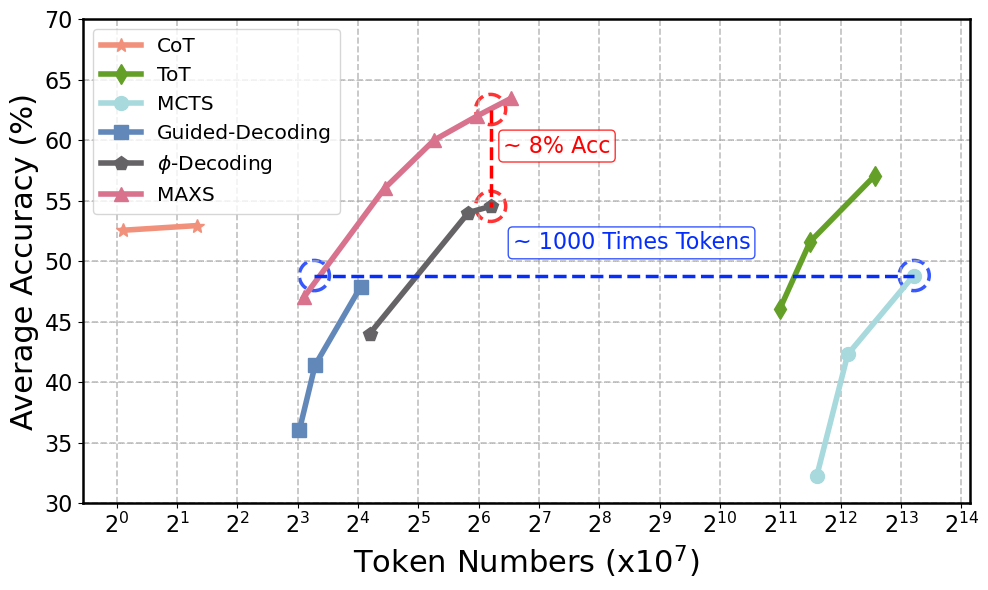} 
\caption{Inference-time scaling law: Accuracy vs. Token usage for different models during decoding.}
\label{fig4}
\end{figure}

\section{Experiments}

\subsection{Experimental Settings}

\paragraph{Benchmarks.} We evaluate our proposed method, MAXS, on five diverse and challenging reasoning benchmarks to assess its performance across both unimodal and multimodal domains. The selected datasets are MathVista, OlympiadBench, TheoremQA, MATH, and EMMA. More dataset details can be found in Appendix~\ref{dataset}.

\paragraph{Backbones and Hyperparameters.} We conduct experiments using three multimodal language models: MiMo-VL-7B, Qwen2.5-VL-7B, and Qwen2.5-VL-32B, to evaluate the robustness and generalizability of MAXS across different architectures and model scales. All experiments are implemented on NVIDIA A800 GPUs with 80GB VRAM, using the vLLM~\cite{kwon2023} inference engine. We keep the decoding configuration fixed for fair comparison, where K = 1, M = 4, and N = 4. Under this setting, the maximum step of reasoning considered is 13. The step scoring strategy is controlled by $\alpha$ = 0.3 and $\beta$ = 0.2, which balance different components of the score. The top-p value is set to 0.95 to ensure a good trade-off between diversity and precision in generation.

\paragraph{Metrics.} We adopt the \textbf{pass@1}~\cite{chen2021evaluatinglargelanguagemodels} rate as our primary accuracy (Acc.) metric to evaluate the correctness of the final generated answer. To measure computational efficiency, we also report the average number of \textbf{input and output tokens} consumed by the backbone model for generating each solution.

\paragraph{Tools.}  
During inference, the LLM agents autonomously invoke external tools to support complex reasoning via code execution and knowledge retrieval. Specifically, a Python-based \textit{Code Interpreter} executes model-generated code for accurate computations, while a \textit{Search Engine} retrieves external knowledge-implemented via an LLM for convenience.

\paragraph{Baselines.}  
We compare MAXS against five representative reasoning methods, including \textit{CoT}, which generates a single step by step reasoning chain, \textit{ToT} and \textit{MCTS}, which explore reasoning trees with pruning via self evaluation or Monte Carlo rollouts, \textit{Guided Decoding}~\cite{NEURIPS2023_81fde95c}, which uses stochastic search guided by self evaluation, and \textit{$\phi$-Decoding}~\cite{xu-etal-2025-ph}, which selects steps based on simulated foresight and path alignment.

\subsection{Main Results}


\begin{figure}[t]
\centering
\includegraphics[width=\columnwidth]{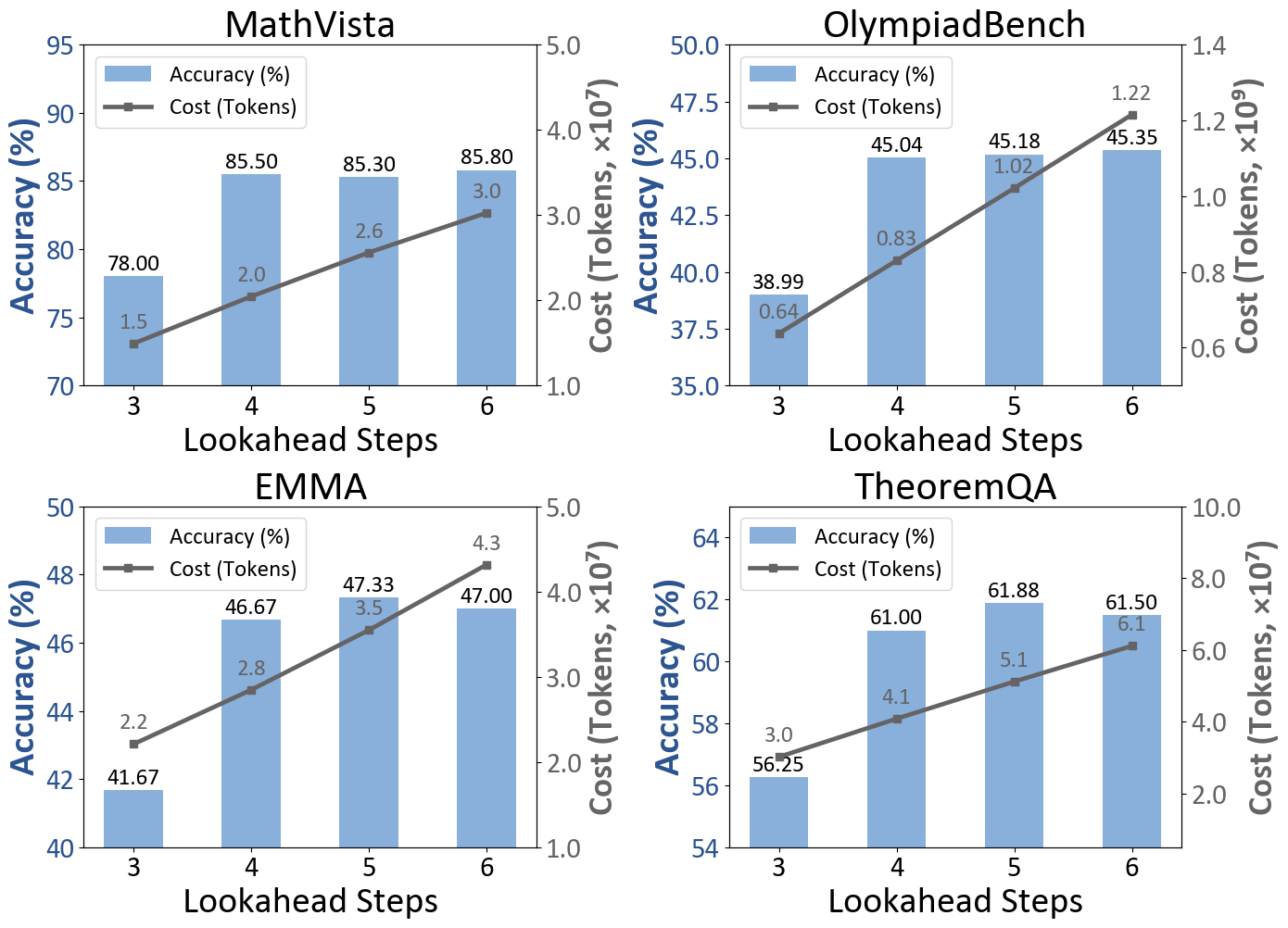}
\caption{Accuracy–cost trade-off under varying lookahead steps across datasets. }
\label{lookahead-step}
\end{figure}

\begin{figure}[ht]
\centering
\includegraphics[width=\columnwidth]{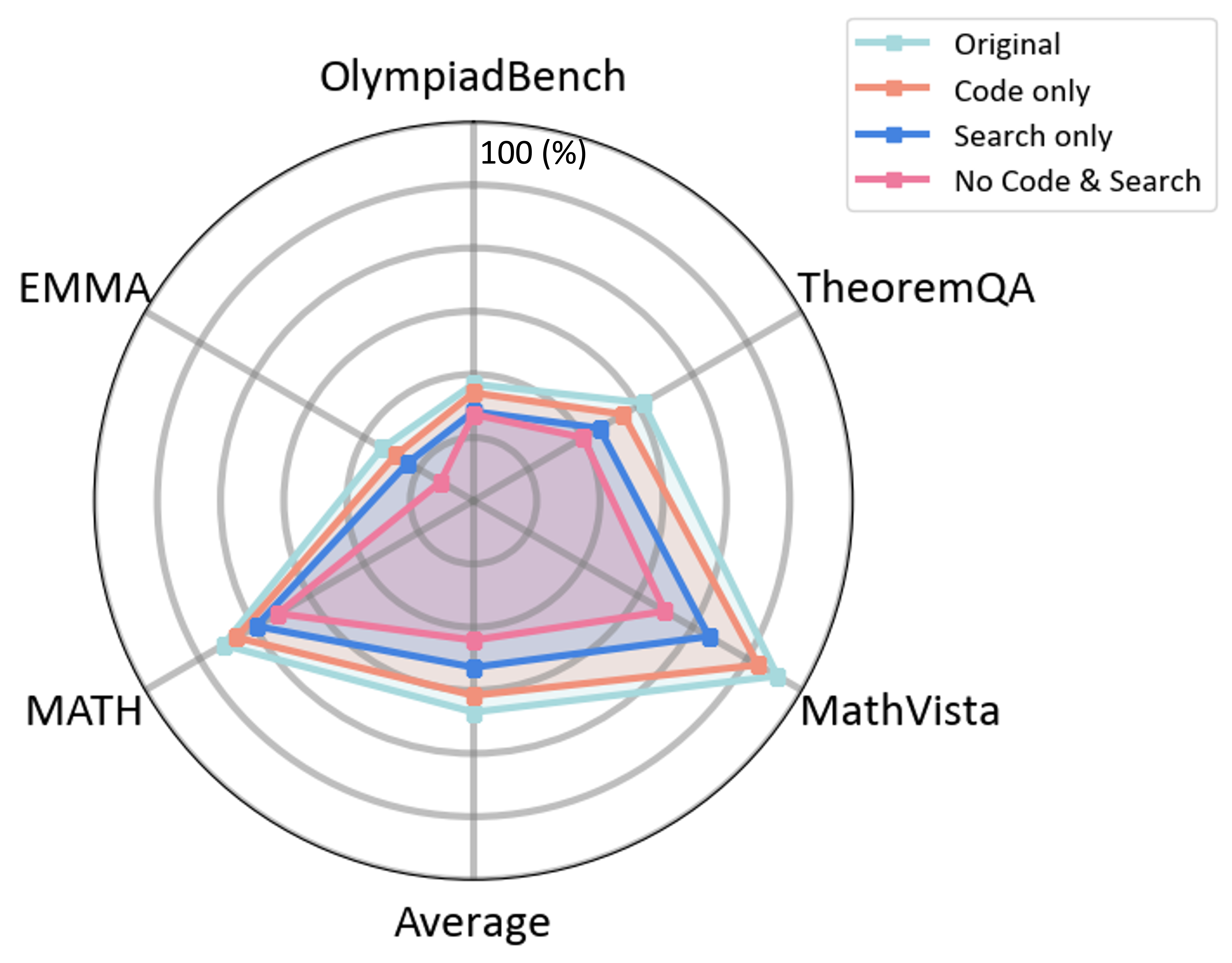}
\caption{Radar plot of accuracy under different tool configurations across datasets.}
\label{tools}
\end{figure}


\paragraph{MAXS improves average performance across backbones.}  
As shown in Table~\ref{table:main}, MAXS consistently outperforms five strong baselines, achieving SOTA results. On MiMo-VL-7B, it reaches 63.46\% accuracy-6.42\% higher than ToT. On Qwen2.5-VL-7B, it surpasses Guided Decoding by 7.43\%, demonstrating strong generalization.

\paragraph{MAXS balances effectiveness and efficiency.}  
While tree-based methods like ToT and MCTS are competitive, they require up to 100× more tokens. On MiMo-VL-7B, MAXS uses $9.86\times10^8$ tokens, compared to ToT's $6.40\times10^{10}$ and MCTS's $9.91\times10^{10}$. Compared to efficient methods like $\phi$-Decoding, MAXS achieves notably higher accuracy with minimal additional cost, reflecting its superior allocation of computation for reasoning.

\subsection{Generalization and Scalability}

\paragraph{MAXS's superiority persists when scaling to the 32B model size.} We conduct experiments on the EMMA benchmark using the Qwen2.5-VL-32B model. 
As shown in Table~\ref{qwen32}, MAXS yields even greater improvements on the larger model, surpassing the strongest baseline, $\phi$-Decoding, by 6.33\%. This confirms its ability to capitalize on the advanced reasoning potential of larger LLMs.


\subsection{Inference-Time Scaling}


\paragraph{MAXS method demonstrates a superior trade-off between performance and computational efficiency.} As shown in Figure~\ref{fig4}, MAXS consistently occupies the optimal top-left region, delivering the highest accuracy for any given token budget on the MiMo-VL-7B model. Horizontally, to achieve a comparable accuracy level of ~49\%, MAXS requires approximately 1,000 times fewer tokens than the MCTS baseline. Vertically, with a similar computational cost to $\phi$-Decoding, MAXS achieves a higher accuracy, showcasing a performance advantage of nearly 8\%.

\section{Analysis}

\subsection{Ablation Studies}

To assess the impact of each component in MAXS, we perform a systematic ablation study by removing one module at a time on MiMo-VL-7B and Qwen2.5-VL-7B. Results in Table~\ref{table:ablation} reveal the following key insights:

\paragraph{Lookahead is essential for globally-aware reasoning.}  
Removing the lookahead module leads to the steepest performance drop (–4.96\% on MiMo-VL, –9.44\% on Qwen2.5-VL), highlighting its role in simulating future trajectories and escaping local optima. This aligns with the Bellman principle and confirms lookahead as fundamental.

\paragraph{Advantage score dominates value estimation.}  
Among the three reward signals, ablating the advantage score yields the greatest degradation, proving it is the key driver of effective step selection. In contrast, step and slope variance mainly aid stability, with smaller impacts.

\paragraph{Trajectory convergence improves efficiency with little cost.}  
Although its removal slightly affects accuracy, trajectory convergence reduces inference cost by terminating redundant rollouts, offering efficiency gains without sacrificing quality.

\begin{figure}[t]
\centering
\includegraphics[width=\columnwidth]{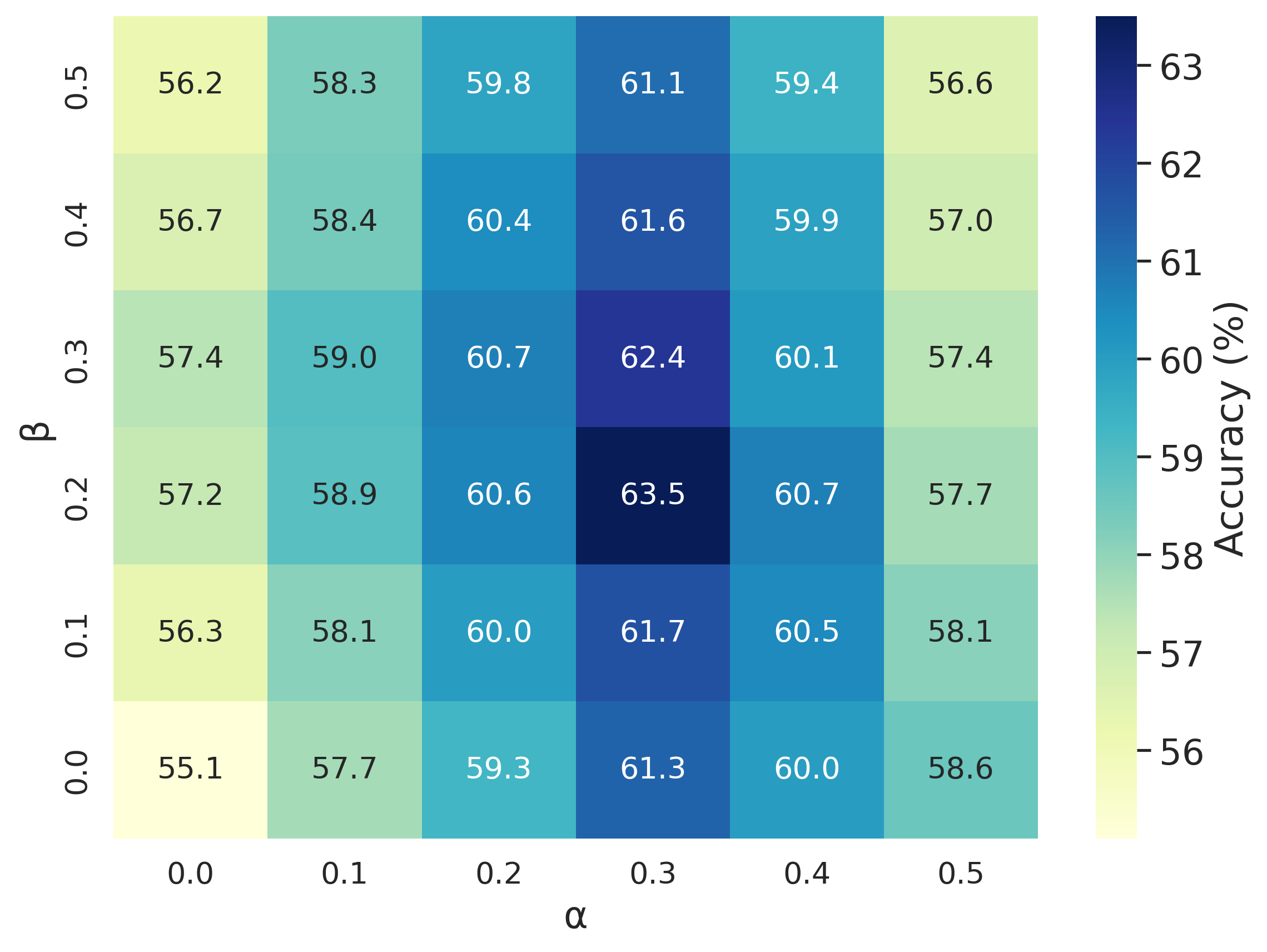}
\caption{Accuracy heatmap under different value estimation weights ($\alpha$, $\beta$) across datasets.}
\label{heatmap}
\end{figure}

\subsection{Analysis of Lookahead Steps}
\paragraph{A 4-step lookahead offers the best balance between accuracy and efficiency.}
As shown in Figure~\ref{lookahead-step}, accuracy improves from 3 to 4 steps but plateaus at 85.3\%–85.8\% beyond that. Meanwhile, token usage rises sharply-from $2.05\times10^7$ at 4-step to $3.07\times10^7$ at 6-step-incurring a 49.8\% overhead. This confirms 4-step as the efficiency frontier, where further gains no longer justify the cost.

\subsection{Analysis of Tool Utilization}


\textbf{Code and search are complementary, removing either harms performance.}
As shown in Figure~\ref{tools}, dropping code or search reduces accuracy from 63.46\% (full model) to 60.81\% (–2.65\%) and 56.36\% (–7.1\%), respectively. The largest drop (52.07\%, –11.4\%) occurs when both are removed, underscoring their synergy in multi-tool reasoning.

\noindent\textbf{Code is especially critical for symbolic reasoning.}
On MathVista, removing code drops accuracy from 85.5\% to 73.0\% (–14.7\%), versus 82.0\% (–4.1\%) without search. While search aids information access, precise computation from code is key to correctness in complex tasks.

\subsection{Analysis of Value Estimation Weights}
\paragraph{Combining step and slope scores ($\alpha{=}0.3$, $\beta{=}0.2$) yields the best overall performance.}
As shown in Figure~\ref{heatmap}, the model achieves peak accuracy (63.5\%) when $\alpha{=}0.3$ and $\beta{=}0.2$, validating the effectiveness of jointly weighting step-based and slope-based rewards in Equation~\ref{eq:joint_reward}. This configuration outperforms the advantage-only baseline ($\alpha{=}0$, $\beta{=}0$, 55.2\%) by +8.3\%. Moreover, adjacent settings also yield competitive results, suggesting that the reward formulation is both robust and well-balanced.

\subsection{Analysis of Reasoning Steps}
\begin{figure}[t]
\centering
\includegraphics[width=\columnwidth]{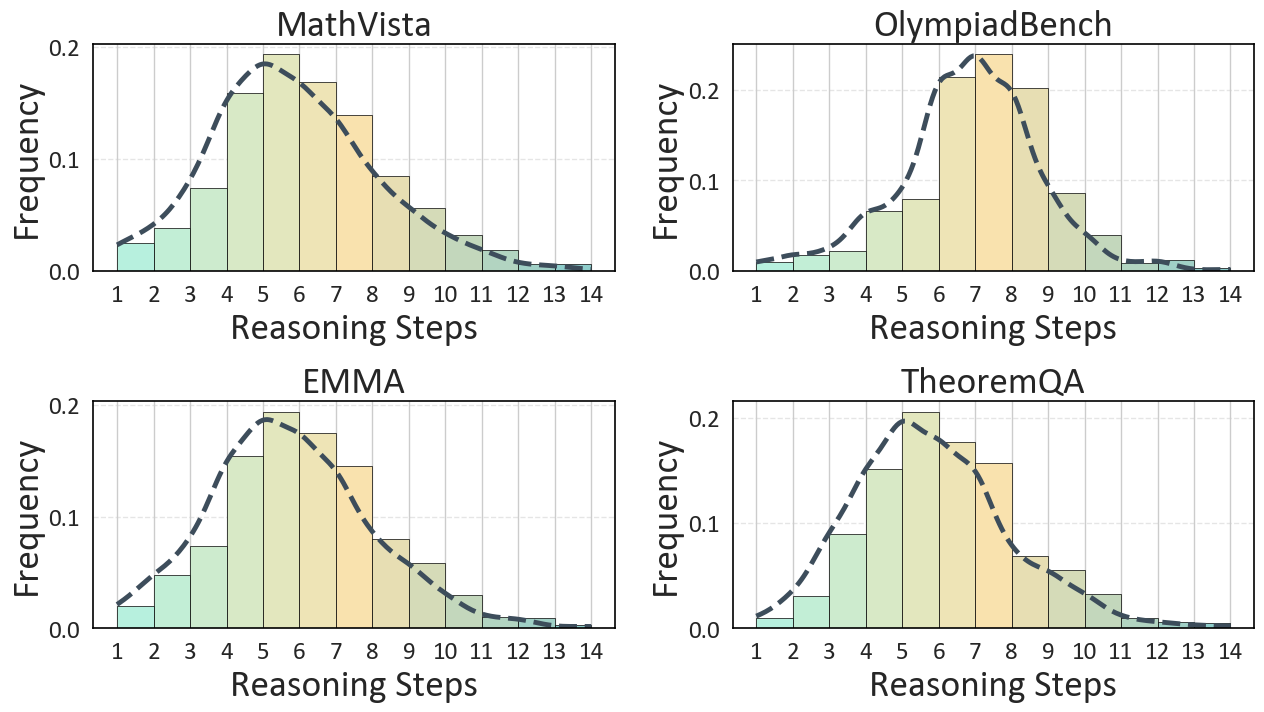}
\caption{Distribution of reasoning steps across datasets.}
\vspace{-0.2cm}
\label{frequency}
\end{figure}

\paragraph{Most problems are solved within 4–8 steps, validating the 13-step cap.}
As shown in Figure~\ref{frequency}, most reasoning trajectories conclude between steps 4 and 8 across datasets. OlympiadBench peaks later at steps 7–8 (23\% each), suggesting greater complexity, while MathVista, EMMA, and TheoremQA concentrate around steps 5–6, covering 58–65\% of cases. Kernel density curves show OlympiadBench spans a broader range (6–9 steps), whereas others are more tightly clustered. Reasoning rarely exceeds 13 steps, justifying our choice of a 13-step cap. These trends confirm that moderate-length trajectories suffice for most problems, with deeper steps reserved for harder cases.

Appendix~\ref{supply} provides additional analysis on rollout, beam size, value estimation methods and significance test, while Appendix~\ref{CaseStudy} presents successful and failure cases.

\section{Related Works}
\vspace{-0.2cm}
\paragraph{LLM Agents and Tool-Augmented Reasoning.}  
LLM Agents enhance language models by dynamically invoking tools (e.g., search, code) to support complex reasoning~\cite{renze2024self, yang2024llm,zhang2026maps,zhang2026mars}. Early approaches insert API calls to improve factual accuracy~\cite{jin2025search, wang2024executable}, while recent frameworks integrate planning and tool selection into multi-step decision-making~\cite{baker2019emergent, torreno2017cooperative, zhang2024survey}. However, most rely on locally greedy decoding and lack long-term tool utility estimation. We address this gap via lookahead-based evaluation and stability-aware step selection.

\paragraph{Inference-Time Scaling and Optimization.}  
Inference-time methods like ToT~\cite{yao2023tree}, MCTS~\cite{gan2025master}, and Best-of-N~\cite{gui2024bonbon} improve answer quality by exploring multiple paths, but often at high computational cost. Efficiency-focused approaches introduce sampling strategies~\cite{manon} with early stopping~\cite{chen2024ee} or pruning~\cite{xu-etal-2025-ph}. Our method complements them by combining lightweight value estimation with convergence-aware rollouts for efficient multi-tool reasoning.

\section{Conclusion}
\vspace{-0.3cm}
In this work, we propose \textit{MAXS}, a meta-adaptive exploration framework that mitigates local myopia and trajectory instability in LLM agents. MAXS integrates lookahead rollouts and a composite value function that incorporates advantage, step variance, and slope variance to guide stable, efficient decision making. A trajectory convergence mechanism further reduces redundant rollouts. Experiments on five benchmarks and three backbones demonstrate improved reasoning performance and reduced cost, with ablations confirming the synergy between lookahead and value-based guidance.

\clearpage

\bibliography{acl_latex}

\clearpage

\appendix

\section{Proof of Proposition}

\subsection{Proof of Proposition 1: Bellman Recursion} \label{A.1}

We aim to prove that the optimal decision at step \( i \) satisfies:
\begin{equation}
\hat{s}_i = \arg\max_{s_i} \left[ R(s_i) + \gamma\, \mathbb{E}_{s_{>i}} V^*(s_{>i}) \right],
\end{equation}
where \( R(s_i) \) is the immediate utility, \( \gamma \in (0,1) \) is a discount factor, and \( V^*(s_{>i}) \) is the expected future value under the optimal policy.

\paragraph{Step 1: Define global optimal value.}
Let the total expected return under the optimal policy starting from the initial input \( s_0 \) be:
\begin{equation}
V^*(s_0) = \max_{s_1, \dots, s_T} \mathbb{E}\left[ \sum_{t=1}^T \gamma^{t-1} R(s_t) \right].
\end{equation}

We can rewrite this recursively as:
\begin{equation}
V^*(s_0) = \max_{s_1} \left[ R(s_1) + \gamma \cdot \mathbb{E}_{s_2} V^*(s_{\geq 2}) \right].
\end{equation}

\paragraph{Step 2: Bellman decomposition at step \( i \).}
At an arbitrary step \( i \), given history \( s_0, \dots, s_{i-1} \), the value function is:
\begin{equation}
\begin{aligned}
V^*(s_{\le i})
&= \max_{s_{>i}} \mathbb{E}\Bigl[
\sum_{k=1}^{K} \gamma^{k-1} R(s_{i+k})
\;\big|\; s_{\le i}
\Bigr],
\end{aligned}
\end{equation}

which can again be written recursively as:
\begin{equation}
\begin{aligned}
V^*(s_{\le i})
&= \max_{s_{i+1}} \Bigl[
R(s_{i+1}) \\
&\qquad + \gamma\, \mathbb{E}_{s_{> i+1}} V^*(s_{> i+1})
\Bigr].
\end{aligned}
\end{equation}

\paragraph{Step 3: Local decision refinement.}
Now consider choosing \( s_i \) to maximize the full downstream return:
\begin{equation}
\begin{aligned}
\hat{s}_i
&= \arg\max_{s_i}\, \mathbb{E}_{s_{>i}} \Bigl[
R(s_i) \\
&\hphantom{= \arg\max_{s_i}\,}\ + \sum_{k=1}^{K} \gamma^{k} R(s_{i+k})
\Bigr].
\end{aligned}
\end{equation}

Let us define:
\begin{equation}
Q(s_i) := R(s_i) + \gamma \cdot \mathbb{E}_{s_{>i}} V^*(s_{>i}),
\end{equation}
then
\begin{equation}
\hat{s}_i = \arg\max_{s_i} Q(s_i).
\end{equation}

\paragraph{Step 4: Relation to lookahead rollout.}
In rollout-based approximation, we generate a set of candidate continuations \( \{s_{>i}^{(k)}\}_{k=1}^M \), then use Monte Carlo estimate:
\begin{equation}
\mathbb{E}_{s_{>i}} V^*(s_{>i}) \approx \frac{1}{M} \sum_{k=1}^M \sum_{j=1}^{K} \gamma^{j-1} R(s_{i+j}^{(k)}),
\end{equation}
which retains consistency with the Bellman optimal formulation.

\paragraph{Conclusion.}
Thus, our decision strategy:
\begin{equation}
\hat{s}_i = \arg\max_{s_i} \left[ R(s_i) + \gamma \cdot \mathbb{E}_{s_{>i}} V^*(s_{>i}) \right]
\end{equation}
recursively links current utility with foresighted trajectory values, consistent with Bellman’s Principle of Optimality.

\subsection{Proof of Proposition 2: Deviation Bound} \label{A.2}

We aim to show that if the step-level variance of a rollout trajectory is bounded by $\varepsilon$, then each individual log-probability score $g_n$ is tightly concentrated around its mean $\bar{g}$:
\begin{equation}
V_{\text{step}} \le \varepsilon \quad \Rightarrow \quad |g_n - \bar{g}| \le \sqrt{N\varepsilon}, 
\end{equation}
$\forall n \in \{1,\dots,N\}.$

\paragraph{Step 1: Definition of variance.}
By definition, the step-level variance of the rollout is:
\begin{equation}
V_{\text{step}} = \frac{1}{N} \sum_{n=1}^{N} (g_n - \bar{g})^2.
\label{eq:a2-def-var}
\end{equation}
This measures the dispersion of log-probabilities across the trajectory.

\paragraph{Step 2: Bounding the $\ell_2$ norm.}
Let $\delta_n := g_n - \bar{g}$ be the deviation from the mean at step $n$. Then:
\begin{equation}
\sum_{n=1}^{N} \delta_n^2 = N \cdot V_{\text{step}} \le N\varepsilon.
\end{equation}
This implies the squared $\ell_2$ norm of the deviation vector $\boldsymbol{\delta} = [\delta_1, \dots, \delta_N]$ is bounded.

\paragraph{Step 3: Derive pointwise bound via inequality.}
Using the fact that:
\begin{equation}
\|\boldsymbol{\delta}\|^2 = \sum_{n=1}^{N} \delta_n^2 \ge \max_n \delta_n^2,
\end{equation}
it follows that for each $n$:
\begin{equation}
|g_n - \bar{g}| = |\delta_n| \le \|\boldsymbol{\delta}\| \le \sqrt{N\varepsilon}.
\end{equation}

\paragraph{Step 4: Alternative probabilistic interpretation.}
Suppose the log-probability sequence $\{g_n\}$ arises from a bounded stochastic process. Then $\bar{g}$ is the empirical mean, and by applying Chebyshev’s inequality:
\begin{equation}
\mathbb{P}(|g_n - \bar{g}| \ge \lambda) \le \frac{V_{\text{step}}}{\lambda^2} \le \frac{\varepsilon}{\lambda^2},
\end{equation}
which shows that the deviation from the mean is highly improbable beyond scale $\sqrt{\varepsilon}$.

\paragraph{Step 5: Connection to discrete Lyapunov stability.}
The result implies that the rollout trajectory is uniformly bounded within a $\sqrt{N\varepsilon}$-ball around the mean, which is a sufficient condition for bounded-input bounded-state (BIBS) stability in discrete-time systems. That is,
$
\forall g_n,\quad |g_n - \bar{g}| \le \mathcal{O}(\sqrt{N\varepsilon}) \quad \Rightarrow \quad \text{bounded trajectory}.
$

\paragraph{Conclusion.}
The variance bound implies that the trajectory exhibits global uniform boundedness, which is analogous to Lyapunov stability in dynamical systems. This supports the interpretation that minimizing $V_{\text{step}}$ leads to smoother and more predictable reasoning behavior.

\subsection{Proof of Proposition 3: Lipschitz Bound} \label{A.3}

We aim to show that if the slope-level variance of the log-probability sequence $\{g_n\}_{n=1}^N$ is bounded by $\varepsilon$, then for any two positions $m,n \in \{1,\dots,N\}$, their cumulative difference is linearly bounded in $|m - n|$:
\begin{equation}
\begin{aligned}
&V_{\text{slope}} \le \varepsilon \\
&\Rightarrow\quad |g_m-g_n| \le \sqrt{(N-1)\varepsilon}\,|m-n|.
\end{aligned}
\end{equation}

\paragraph{Step 1: Define local slope sequence.}
Let $\delta_n := g_{n+1} - g_n$ be the first-order discrete derivative (slope) between adjacent log-probability values:
\begin{equation}
\delta_n = g_{n+1} - g_n, \quad \text{for } n = 1,\dots,N-1.
\end{equation}

Let the average slope be:
\begin{equation}
\bar{\delta} = \frac{1}{N-1} \sum_{n=1}^{N-1} \delta_n.
\end{equation}

\paragraph{Step 2: Define slope-level variance.}
The slope variance is defined as:
\begin{equation}
V_{\text{slope}} = \frac{1}{N-1} \sum_{n=1}^{N-1} (\delta_n - \bar{\delta})^2.
\label{eq:slope-var}
\end{equation}
This measures the local fluctuation in directional progress. Let $\Delta_n := \delta_n - \bar{\delta}$ denote the deviation from average slope.

Then,
\begin{equation}
\sum_{n=1}^{N-1} \Delta_n^2 = (N-1)\cdot V_{\text{slope}} \le (N-1)\varepsilon.
\end{equation}

\paragraph{Step 3: Express global difference via telescoping sum.}
Let $m < n$ without loss of generality. Then we have:
\begin{equation}
g_n - g_m = \sum_{k=m}^{n-1} \delta_k = (n - m)\bar{\delta} + \sum_{k=m}^{n-1} \Delta_k.
\label{eq:gn-gm-decompose}
\end{equation}
The first term captures the trend, and the second term reflects local irregularity.

\paragraph{Step 4: Bound the deviation term.}
By Cauchy–Schwarz inequality:
\begin{align}
\left| \sum_{k=m}^{n-1} \Delta_k \right|^2 &\le (n - m) \cdot \sum_{k=m}^{n-1} \Delta_k^2 \\
&\le (n - m) \cdot \sum_{k=1}^{N-1} \Delta_k^2 \\
&\le (n - m)(N-1)\varepsilon.
\end{align}
Hence,
\begin{equation}
\left| \sum_{k=m}^{n-1} \Delta_k \right| \le \sqrt{(n - m)(N - 1)\varepsilon}.
\end{equation}

\paragraph{Step 5: Final bound on log-probability difference.}
From Eq.~\eqref{eq:gn-gm-decompose}, we have:
\begin{equation}
\begin{aligned}
|g_n-g_m|
&\le |n-m|\,|\bar{\delta}| \\
&\quad + \sqrt{(n-m)(N-1)\varepsilon}.
\end{aligned}
\end{equation}

In worst-case or centered-slope settings (e.g., $\bar{\delta} \approx 0$), the term simplifies to:
\begin{equation}
|g_n - g_m| \le \sqrt{(N - 1)\varepsilon} \cdot |n - m|,
\end{equation}
which mimics the discrete Lipschitz condition with constant $\sqrt{(N - 1)\varepsilon}$.

\paragraph{Step 6: Discrete Lipschitz analogy.}
A function $f(x)$ is Lipschitz continuous if:
\begin{equation}
|f(x) - f(y)| \le L |x - y|, \quad \forall x, y.
\end{equation}
Here, the sequence $\{g_n\}$ exhibits analogous behavior, where the bounded variance on discrete slopes constrains global oscillation across the trajectory.

\paragraph{Conclusion.}
The slope variance $V_{\text{slope}}$ directly governs the rate of directional fluctuation. Bounding it enforces path regularity, controls local curvature, and promotes globally smooth reasoning progress. This justifies the slope-consistency reward in our value function as a surrogate for discrete Lipschitz continuity.

\section{Datasets} \label{dataset}

As illustrated in Table~\ref{datasets_details}, this study utilizes five publicly available datasets: \textit{MathVista}, \textit{OlympiadBench}, \textit{EMMA}, \textit{TheoremQA}, and \textit{MATH}. These benchmarks cover a wide range of science problems and are widely used for evaluating reasoning abilities of large language models.

\begin{table}[t]
\centering
\resizebox{\columnwidth}{!}{
\begin{tabular}{lll}
\toprule
\textbf{Dataset} & \textbf{Category} & \textbf{Size} \\
\midrule
\textbf{MathVista} & Overall & 1000 \\
\midrule
\multirow{10}{*}{\textbf{OlympiadBench}} 
& OE\_TO\_maths\_zh\_CEE & 1240 \\
& OE\_MM\_maths\_zh\_CEE & 1910 \\
& OE\_TO\_physics\_en\_COMP & 236 \\
& OE\_MM\_maths\_en\_COMP & 150 \\
& OE\_MM\_physics\_en\_COMP & 456 \\
& OE\_TO\_maths\_en\_COMP & 674 \\
& OE\_TO\_maths\_zh\_COMP & 408 \\
& OE\_MM\_physics\_zh\_CEE & 1483 \\
& OE\_MM\_maths\_zh\_COMP & 56 \\
& OE\_TO\_physics\_zh\_CEE & 115 \\
\cmidrule{2-3}
& maths (subset total) & 4438 \\
& physics (subset total) & 2290 \\
\cmidrule{2-3}
& Overall & 6728\\
\midrule
\multirow{3}{*}{\textbf{EMMA}} 
& Math & 100 \\
& Physics & 100 \\
& Chemistry & 100 \\
\cmidrule{2-3}
& Overall & 300\\
\midrule
\textbf{TheoremQA} & Overall & 800 \\
\midrule
\textbf{MATH} & Sampled & 300 \\
\bottomrule
\end{tabular}}
\caption{Detailed composition of the five datasets used in our study: MathVista, OlympiadBench, EMMA, TheoremQA, and MATH. For OlympiadBench, we present its fine-grained subsets along with their corresponding sizes. We also report the total number of problems in the math- and physics-related subsets, where applicable. For EMMA, we adopt its MINI version, and for MATH, we sample 300 problems from the full dataset.}
\label{datasets_details}
\end{table}

\paragraph{MathVista.} MathVista is a large-scale scientific reasoning dataset that spans multiple reasoning types such as algebraic, geometric, statistical, scientific, numeric commonsense, and logical reasoning, aiming to assess the comprehensive capabilities of machine learning models in solving complex scientific problems. The dataset(\textit{testmini}) contains 1,000 data points covering various issues across multiple disciplines, designed with varying difficulty levels to help researchers evaluate model reasoning abilities. The release of MathVista supports interdisciplinary scientific research.

\paragraph{OlympiadBench.} OlympiadBench consists of two subdomains, maths and physics, and is specifically designed for Mathematical and Physical Olympiads, featuring a wide range of challenging problems to assess models' performance on high-level scientific tasks. The dataset contains two difficulty levels: competition level and college level, reflecting the diversity and depth of real-world Olympiad problems. It includes two types of questions: open-ended questions and theorem-proof questions. To focus on evaluating generative mathematical reasoning abilities, we select the 6,728 open-ended(OE) questions for our experiments.

\paragraph{EMMA.} EMMA is a multimodal scientific reasoning dataset covering three subsets: Math, Physics, and Chemistry. By integrating mathematical expressions, physical formulas, and chemical symbols with natural language descriptions, it focuses on testing models' abilities in interdisciplinary scientific reasoning. This version uses the EMMA dataset, which contains 100 data points from each subdomain (mathematics, physics, and chemistry).

\paragraph{TheoremQA.} TheoremQA is a benchmark dataset designed to evaluate the ability of language models to perform theorem-based reasoning. It contains 800 high-quality question-answer pairs grounded in over 350 unique theorems, covering fields such as mathematics, physics, electrical engineering, computer science, and finance. The dataset focuses on assessing whether models can correctly apply formal theorems to solve advanced problems, making it a valuable resource for studying scientific reasoning in large language models.

\paragraph{MATH.} MATH is a benchmark dataset designed to evaluate the advanced mathematical reasoning capabilities of language models. It comprises 12,500 high school competition-level problems drawn from sources such as AMC, AIME, and other standardized exams. The dataset spans seven mathematical domains: Prealgebra, Algebra, Number Theory, Counting \& Probability, Geometry, Intermediate Algebra, and Precalculus. Each problem includes a detailed step-by-step solution, final answer, subject label, and difficulty rating, allowing for fine-grained analysis of model performance across diverse mathematical topics. We randomly sampled 300 problems from the MATH dataset, selecting 60 problems from each difficulty level (Levels 1 through 5) to ensure an evenly balanced coverage across difficulty tiers.

\section{MAXS Decoding Algorithm} \label{decoding}

\begin{algorithm}[tb]
\caption{MAXS Decoding with Lookahead and Value Estimation}
\label{alg:maxs}
\textbf{Input}: Input prompt $s_0$\\
\textbf{Parameter}: Model $\pi_\theta$, beam size $K$, temperature $\tau$, threshold $\delta$, rollout size $M$, lookahead size $N$\\
\textbf{Output}: Final reasoning trajectory $s = \{s_1, \dots, s_T\}$

\begin{algorithmic}[1]
\STATE Initialize $t \gets 1$, $s \gets \{s_0\}$
\WHILE{not end-of-sequence}
    \STATE Sample $K$ candidates $\{s_t^{(m)}\}_{m=1}^M \sim \pi_\theta(s_t \mid s_{<t})$
    \FOR{each candidate $s_t^{(m)}$}
        \STATE Rollout $s_{>t}^{(m)} \sim \pi_\theta$ up to length $N$
        \STATE Compute foresight $F_t^{(k)} = \pi_\theta(s_{>t}^{(k)} \mid s_{\le t}^{(k)})$
        \STATE Compute advantage $R^{\text{adv}}_t$, step variance $R^{\text{step}}_t$, slope variance $R^{\text{slope}}_t$
        \STATE Aggregate reward $R^{(k)}$ via Eq.~\eqref{eq:joint_reward}
    \ENDFOR
    \IF{$\text{Var}(\{R^{(k)}\}) \le \delta$}
        \STATE Break rollout, continue auto-regressive decoding
    \ENDIF
    \STATE Select $\hat{s}_t \sim \text{softmax}(R^{(k)} / \tau)$
    \STATE Append $\hat{s}_t$ to $s$, update $t \gets t+1$
\ENDWHILE
\STATE \textbf{return} sequence $s$
\end{algorithmic}
\end{algorithm}

We summarize the full decoding process in Algorithm~\ref{alg:maxs}. At each step $t$, the model samples $K$ candidate actions $\{s_t^{(k)}\}_{k=1}^K$ from the policy $\pi_\theta$. For each candidate, a stochastic rollout generates future steps $s_{>t}^{(k)}$, from which the foresight probability $F_t^{(k)}$ is estimated.

We compute the composite reward $R^{(k)}$ using advantage score, step-level variance, and slope-level variance, combined via Eq.~\eqref{eq:joint_reward}. If the reward variance $\text{Var}(\{R^{(k)}\})$ falls below threshold $\delta$, we terminate rollout early and resume auto-regressive decoding.
Otherwise, the next step $\hat{s}_t$ is sampled according to $\text{softmax}(R^{(k)} / \tau)$ and appended to the sequence. This process iterates until an end-of-sequence token is generated.

\section{Supplement Analysis} \label{supply}
\subsection{Analysis of Rollout Steps}

\begin{figure}[t]
\centering
\includegraphics[width=\columnwidth]{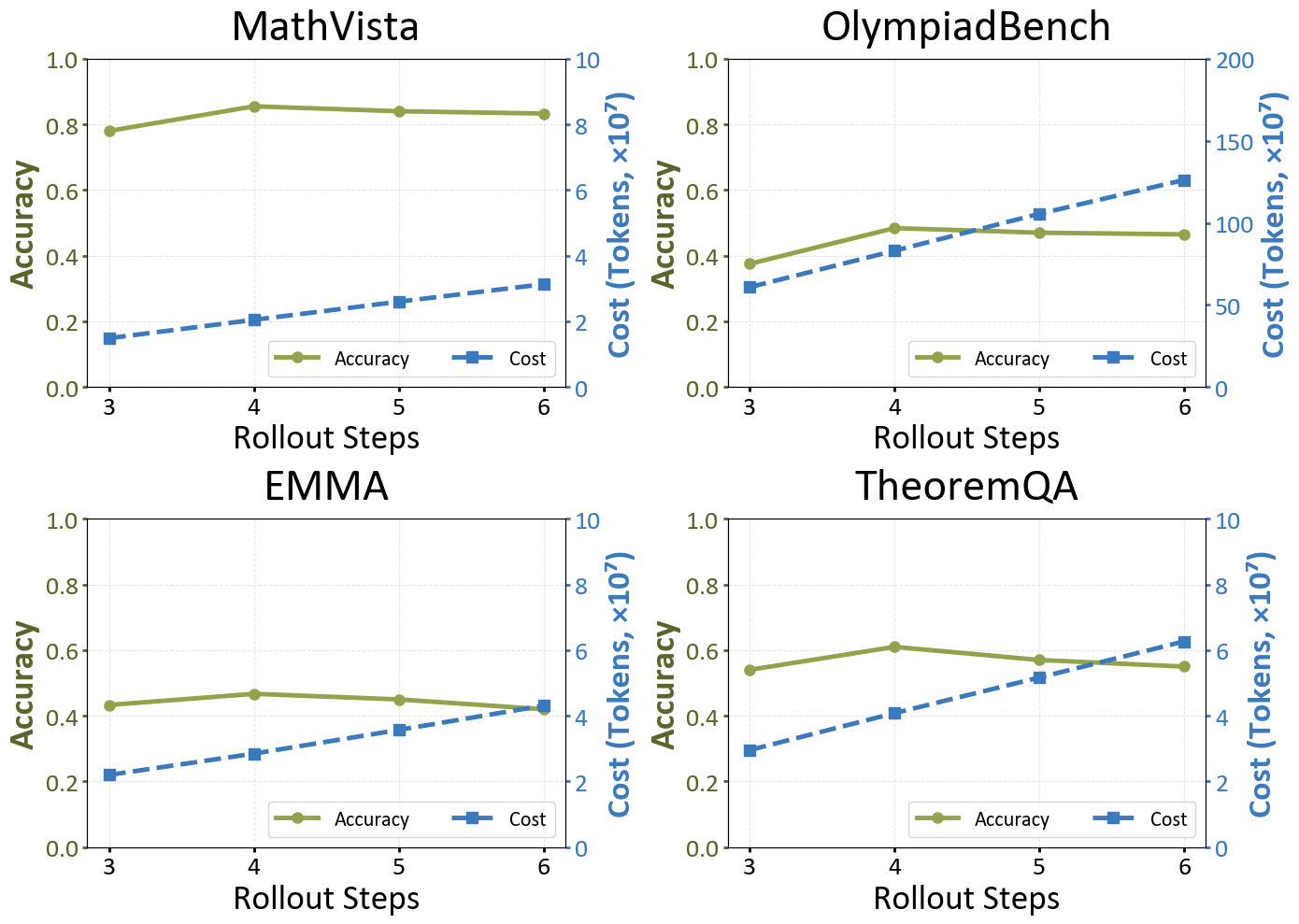}
\caption{Accuracy–cost trade-off under varying rollout steps across datasets.}
\label{rollout-step}
\end{figure}

\begin{figure}[t]
\centering
\includegraphics[width=\columnwidth]{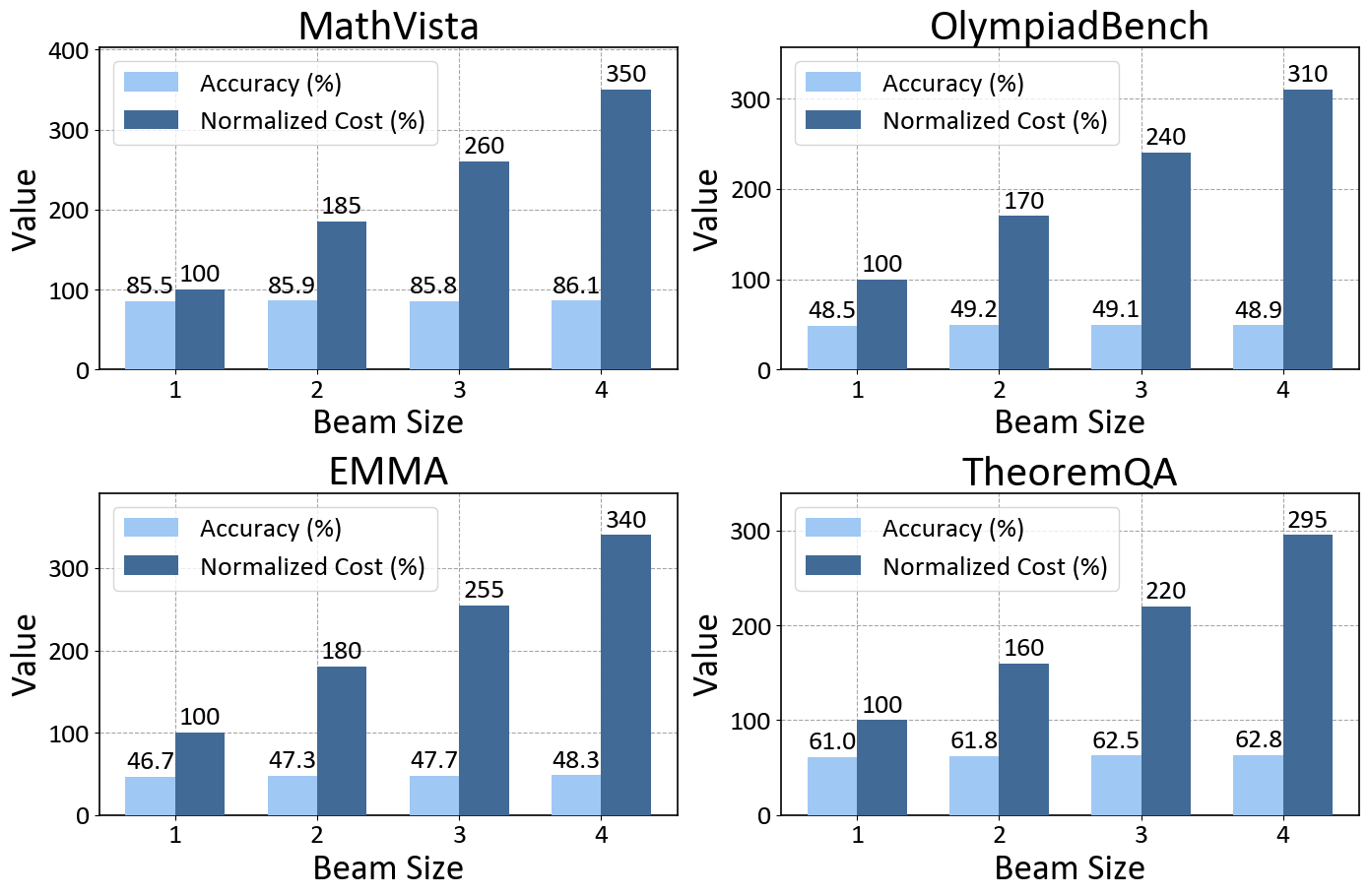}
\caption{Accuracy vs. relative cost under varying beam sizes (1-beam normalized to 100\%).}
\label{beam-size}
\end{figure}

\begin{figure*}[ht]
\centering
\includegraphics[width=0.8\textwidth]{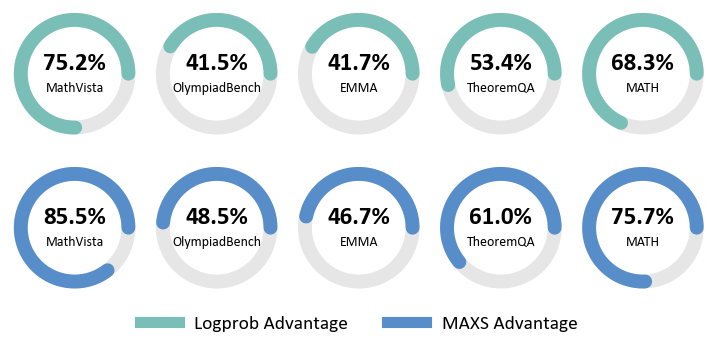}
\caption{Comparison of different value estimation methods across datasets.}
\label{advantage}
\end{figure*}

\paragraph{Rollout steps beyond 4 incur excessive cost with no accuracy gain.}
As shown in Figure~\ref{rollout-step}, accuracy on OlympiadBench improves from 0.375 to 0.484 when increasing the rollout steps from 3 to 4, but declines thereafter. Meanwhile, token cost rises sharply-from 332M at 3-step to 564M at 5-step and 661M at 6-step. This confirms 4-step as the efficiency frontier, where further rollout yields diminishing or even negative returns.

\subsection{Analysis of Beam Size}
\paragraph{1-beam strikes the best balance between accuracy and cost.}
Figure~\ref{beam-size} shows that 1-beam maintains normalized computational cost at 100\% (leftmost dark blue bars). Increasing to 4-beam dramatically raises costs-by +250\% on MathVista, +195\% on TheoremQA, and +180\% on EMMA-while accuracy gains remain marginal ($<1.5\%$). On OlympiadBench, accuracy rises by only 0.46\% despite a 210\% cost increase. These results confirm that larger beams yield diminishing returns, with 1-beam offering the most efficient trade-off.

\subsection{Comparison of Value Estimation Methods}
\textbf{MAXS consistently outperforms Logprob–based value estimation.}
As shown in Figure~\ref{advantage}, MAXS achieves 5.0–10.3\% higher accuracy across all five reasoning benchmarks, with the largest gains observed on MathVista and TheoremQA. This confirms our value estimation method's superiority in modeling complex reasoning trajectories, especially in symbolic tasks where log-probability fails to capture structural value. The stable margin of 5.0–7.3\% on OlympiadBench, EMMA, and MATH further demonstrates MAXS’s robustness across diverse reasoning formats.

\begin{table}[h]
    \centering
    \small
    \resizebox{\columnwidth}{!}{
    \begin{tabular}{lcc}
        \toprule
        \textbf{Comparison} & \textbf{$p$-value} & \textbf{Significance} \\
        \midrule
        \multicolumn{3}{c}{\textbf{MiMo-VL-7B-SFT}} \\
        \midrule
        MAXS vs. CoT              & $< 0.001$  & \checkmark \\
        MAXS vs. ToT              & $< 0.001$  & \checkmark \\
        MAXS vs. MCTS             & $< 0.001$  & \checkmark \\
        MAXS vs. Guided Decoding  & $< 0.001$  & \checkmark \\
        MAXS vs. $\phi$-Decoding  & $< 0.001$  & \checkmark \\
        \midrule
        \multicolumn{3}{c}{\textbf{Qwen2.5-VL-7B-Instruct}} \\
        \midrule
        MAXS vs. CoT              & $< 0.001$  & \checkmark \\
        MAXS vs. ToT              & $< 0.001$  & \checkmark \\
        MAXS vs. MCTS             & $< 0.001$  & \checkmark \\
        MAXS vs. Guided Decoding  & $< 0.001$  & \checkmark \\
        MAXS vs. $\phi$-Decoding  & $< 0.001$  & \checkmark \\
        \bottomrule
    \end{tabular}}
    \caption{Results of McNemar’s Test for Statistical Significance. We compare our proposed MAXS method against all baseline methods across two base models. A $p$-value $< 0.05$ indicates a statistically significant difference. As shown, MAXS demonstrates significant improvement over all baselines.}
    \label{tab:significance_test_models}
\end{table}

\subsection{Significance Test}

To determine whether the gains achieved by MAXS are statistically significant, we perform McNemar’s test for paired comparisons between MAXS and each baseline method. Table~\ref{tab:significance_test_models} reports the results on two backbones, MiMo-VL-7B-SFT and Qwen2.5-VL-7B-Instruct. Across all comparisons, including strong baselines such as ToT and $\phi$-Decoding, MAXS achieves $p<0.001$, which is well below the significance threshold $\alpha=0.05$. These results indicate that the improvements of MAXS over existing decoding strategies are statistically significant and consistent across model architectures.

\section{Case Study}\label{CaseStudy}

In this section, we present a successful case (Figure~\ref{case-study}) and a failure case (Figure~\ref{failurecase}), respectively.

\subsection{Successful Case}
\begin{figure*}[t]
	\large
	\centering
	\includegraphics[scale=0.32]{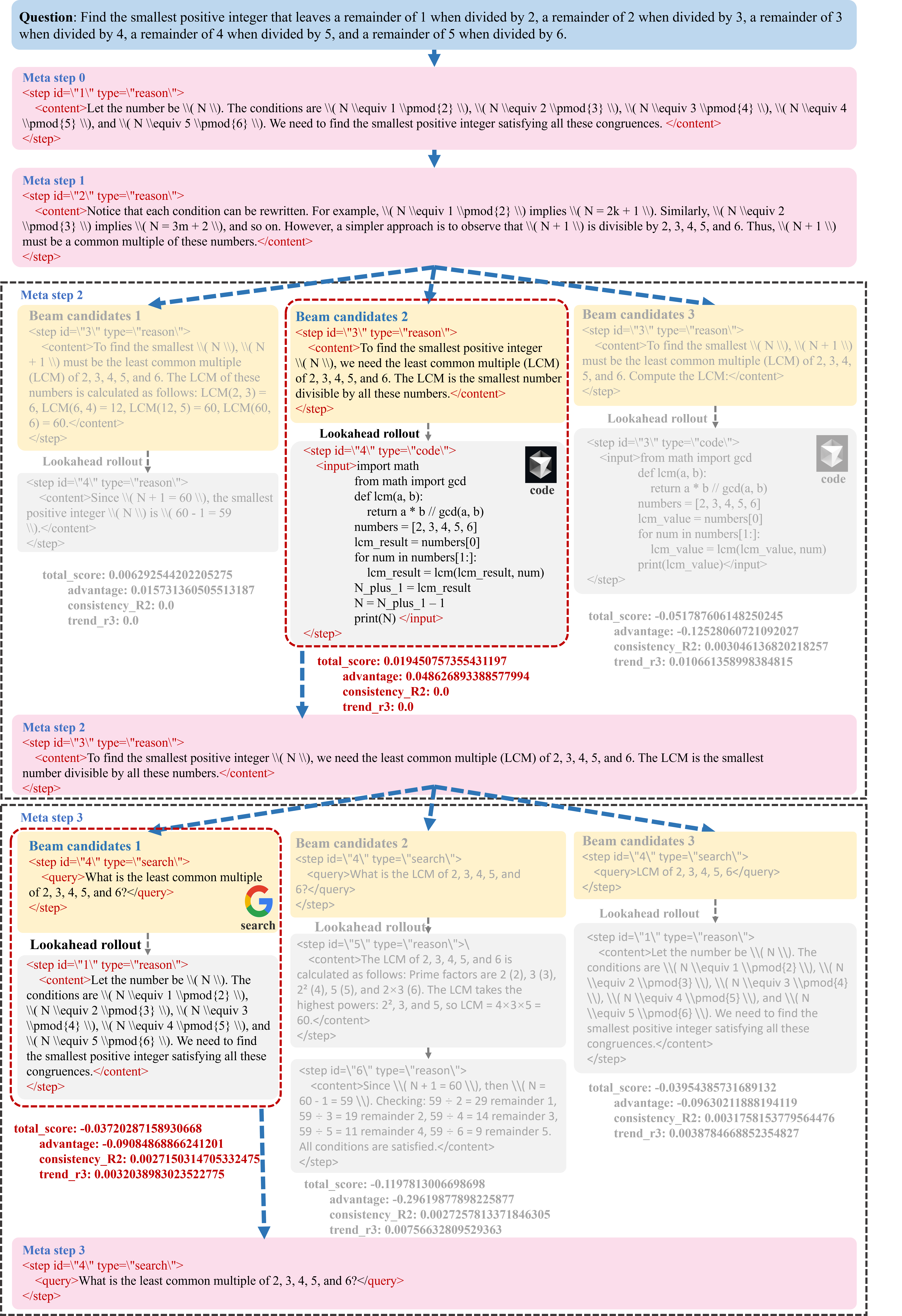}
	\caption{Successful case of MAXS solving a TheoremQA problem. At each step, it performs rollout and foresight (up to four steps), evaluates candidates via three advantage metrics, and iteratively selects the best path. The process dynamically integrates reasoning, search, and tool use.}
	\label{case-study}
\end{figure*}

Figure~\ref{case-study} presents an example of problem-solving using the MAXS method, with the question sourced from the TheoremQA dataset. As shown in steps 2 and 3, MAXS performs a rollout at each reasoning step, exploring multiple candidate reasoning paths. After generating beam candidates, the model conducts foresight for each path. Although the foresight depth is set to 4, in later stages of the reasoning process, the solution may be completed within fewer than four steps-thus not every step features a full four-step foresight chain. Following this, MAXS evaluates each rollout plus foresight chain using the three advantage metrics proposed in this paper (Advantage Score, Step-Level Variance, and Slope-Level Variance) and selects the candidate with the highest overall score as the action for the current step. This process continues iteratively until the final solution is reached. Notably, each candidate or foresight step may involve different types of operations such as reasoning, search, or code execution. The model dynamically invokes external tools to ensure high-quality reasoning throughout the problem-solving process.

\subsection{Failure Case}

\begin{figure*}[t]
	\large
	\centering
	\includegraphics[scale=0.32]{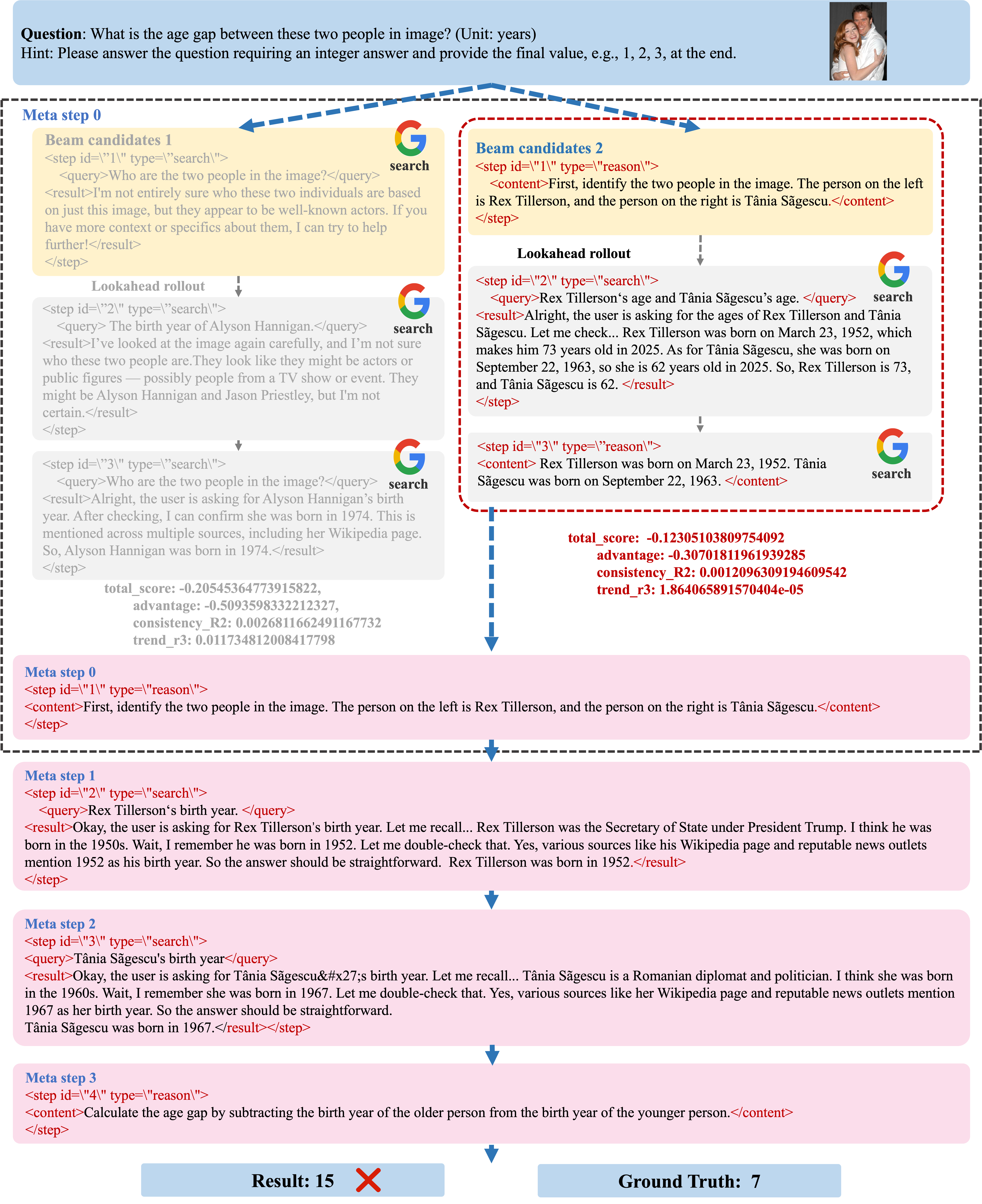}
	\caption{A failure case on the MathVista dataset where MAXS selects an incorrect visual recognition path due to the low confidence of search tool results. The initial misidentification of the individuals propagates through the reasoning chain, leading to an erroneous final answer despite valid subsequent calculations.}
	\label{failurecase}
\end{figure*}

Figure~\ref{failurecase} presents a failure case of MAXS on MathVista, illustrating how an early recognition error can derail multi-step reasoning. The task asks for the age difference between two individuals shown in an image. At the initial stage (Meta step~0), MAXS performs a rollout and generates two beam candidates. Beam~1 attempts to use the search tool to identify the individuals, but the returned results are ambiguous and do not yield a reliable match, leading to low confidence and a lower evaluation score ($-0.205$). Beam~2 instead relies on the model's internal visual recognition. Although it misidentifies the individuals as Rex Tillerson and T\^{a}nia S\k{a}gescu, it produces a coherent explanation and receives a higher score ($-0.123$). MAXS therefore selects Beam~2 and commits to an incorrect premise.

This initial mistake propagates through later steps. In Meta steps~1-3, the model retrieves birth information for the misidentified subjects and performs the arithmetic correctly, but the final answer is necessarily wrong: it outputs 15 years instead of the ground-truth 7 years. This case highlights a limitation of the system: when tool-based retrieval is uncertain or ineffective, the model may prefer a more confident but incorrect internal hypothesis, which can dominate the downstream reasoning process.

\end{document}